\documentclass[11pt]{article}

\usepackage[letterpaper, margin=1in]{geometry}
\usepackage{amsmath}
\usepackage{amssymb}
\usepackage{bm}
\usepackage{enumitem}
\usepackage{float}
\usepackage{graphicx}
\usepackage{xcolor}

\usepackage{hyperref}

\usepackage{ifthen}
\newif\ifshortver
\usepackage{caption}
\usepackage{subcaption}

\newcommand{\ifshort}[2]{\ifshortver#1\else#2\fi}

\PassOptionsToPackage{sort, compress}{natbib}
\usepackage{natbib}

\usepackage{Shorthands}
\usepackage[utf8]{inputenc} 
\usepackage[T1]{fontenc}    
\usepackage{hyperref}       
\usepackage{url}            
\usepackage{booktabs}       
\usepackage{amsfonts}       
\usepackage{nicefrac}       
\usepackage{microtype}      
\usepackage{xcolor}         
\usepackage[font=small,labelfont=bf]{caption}               
\usepackage{wrapfig}
\usepackage{float}
\usepackage{algorithm}
\usepackage{algpseudocode}
\usepackage{thmtools, thm-restate}

\hypersetup{
    unicode=false,          
    pdftoolbar=true,        
    pdfmenubar=true,        
    pdffitwindow=false,     
    pdfstartview={FitH},    
    pdfnewwindow=true,      
    colorlinks=true,       
    linkcolor=blue,          
    citecolor=black,        
    filecolor=magenta,      
    urlcolor=blue,           
    breaklinks=true
}

\title{Sample-Efficient Linear Representation Learning from Non-IID Non-Isotropic Data}

\author{Thomas T.\ Zhang$^1$\footnote{Corresponding author, email: \texttt{ttz2@seas.upenn.edu}}\;, Leonardo F.\ Toso$^2$, James Anderson$^2$, Nikolai Matni$^1$}
\date{$^1$ Department of Electrical and Systems Engineering, University of Pennsylvania \\
$^2$ Department of Electrical Engineering, Columbia University}

\begin{document}

\maketitle

\vspace{-0.4cm}

\begin{abstract}
A powerful concept behind much of the recent progress in machine learning is the extraction of common features across data from heterogeneous sources or tasks. Intuitively, using all of one's data to learn a common representation function benefits both computational effort and statistical generalization by leaving a smaller number of parameters to fine-tune on a given task. Toward theoretically grounding these merits, we propose a general setting of recovering linear operators $M$
from noisy vector measurements $y = Mx + w$, where the covariates $x$ may be both non-i.i.d.\ and non-isotropic. We demonstrate that existing isotropy-agnostic representation learning approaches incur biases on the representation update, which causes the scaling of the noise terms to lose favorable dependence on the number of source tasks. This in turn can cause the sample complexity of representation learning to be bottlenecked by the single-task data size. We introduce an adaptation, \texttt{De-bias \& Feature-Whiten} (\texttt{DFW}), of the popular alternating minimization-descent scheme proposed independently in \ifshort{Collins et al., (2021) \citep{collins2021exploiting}}{\cite{collins2021exploiting}} and \cite{nayer2022fast}, and establish linear convergence to the optimal representation with noise level scaling down with the \emph{total} source data size. This leads to generalization bounds on the same order as an oracle empirical risk minimizer. We verify the vital importance of \texttt{DFW} on various numerical simulations. In particular, we show that vanilla alternating-minimization descent fails catastrophically even for iid, but mildly non-isotropic data.
Our analysis unifies and generalizes prior work, and provides a flexible framework for a wider range of applications, such as in controls and dynamical systems. 





\end{abstract}

\section{Introduction}


\sloppy
A unifying paradigm belying recent exciting progress in machine learning is learning a common feature space or \emph{representation} for downstream tasks from heterogeneous sources. This forms the core of fields such as meta-learning, transfer learning, and federated learning. A shared theme across these fields is the scarcity of data for a specific task out of many, such that designing individual models for each task is both computationally and statistically inefficient, impractical, or impossible. Under the assumption that these tasks are similar in some way, a natural alternative approach is to use data across many tasks to learn a common component, such that fine-tuning to a given task involves fitting a much smaller model that acts on the common component. Over the last few years, significant attention has been given to framing this problem setting theoretically, providing provable benefits of learning over multiple tasks in the context of linear regression \citep{bullins2019generalize, du2020few, tripuraneni2021provable, collins2021exploiting, thekumparampil2021sample, saunshi2021representation} and in identification/control of linear dynamical systems \citep{modi2021joint, chen2023multi, zhang2022multitask}. These works study the problem of \emph{linear representation learning}, where the data for each task is generated noisily from an unknown shared latent subspace, and the goal is to efficiently recover a representation of the latent space $\hat\Phi$ from data across different task distributions. For example, in the linear regression setting, one may have data of the form
\[
\yhi = {\theta^{(t)}}^\top \Phi \xhi + \mathrm{ noise}, \quad \yhi \in \R, \xhi \in \R^{d_x}, \Phi \in \R^{r \times d_x},
\]
with $i=1,\dots, N$ iid data points from $t=1,\dots, T$ task distributions. Since the representation $\Phi$ is shared across all tasks, one may expect the generalization error of an approximate representation $\hat\Phi$ fit on $TN$ data points to scale as $\frac{d_x r}{TN}$, where $d_x r$ is the number of parameters determining the representation. This is indeed the flavor of statistical guarantees from prior work \citep{du2020few, tripuraneni2021provable, thekumparampil2021sample, zhang2022multitask}, which concretely demonstrates the benefit of using data across different tasks.

However, existing work, especially beyond the scalar measurement setting, is limited in one or more important components of their analysis.  For example, it is common to assume that the covariates  $\xhi$ are isotropic across all tasks. Furthermore, statistical analyses often assume access to an empirical risk minimizer, even though the linear representation learning problem is non-convex and ill-posed \citep{maurer2016benefit, du2020few, zhang2022multitask}.
Our paper addresses these problems under a unified framework of \textit{linear operator recovery}, i.e.\ recovering linear operators $M \in \R^{d_y \times d_x}$ from (noisy) vector measurements $y = Mx + w$, where the covariates $x$ may not be independent or isotropic. This setting subsumes the scalar measurement setting, and encompasses many fundamental control and dynamical systems problems, such as linear system identification and imitation learning. 
In particular, the data in these settings are incompatible with the common distributional assumptions (e.g., independence, isotropy) made in prior work. 


\paragraph{Contributions:} Toward this end, our main contributions are as follows: 
\vspace{-.5em}
\begin{itemize}[left=0pt, noitemsep]

    \item
    We demonstrate that naive implementation of local methods for linear representation learning fail catastrophically even when the data is iid but mildly non-isotropic. We identify the source of the failure as interaction between terms incurring biases in the representation gradient, which do not scale down with the number of tasks.

    \item We address these issues by introducing two practical algorithmic adjustments, \texttt{De-bias \& Feature-Whiten} (\texttt{DFW}), which provably mitigate the identified issues. We then show that \texttt{DFW} is necessary for gradient-based methods to benefit from the total size of the source dataset.

    
    
    
    \item We numerically show our theoretical guarantees are predictive of the efficacy of our proposed algorithm, and of the key importance of individual aspects of our algorithmic framework.

\end{itemize}

Our main result can be summarized by the following descent guarantee for our proposed algorithm.
\begin{theorem}[main result, informal] Let $\hat\Phi$ be the current estimate of the representation, and $\Phi_\star$ the optimal representation. Running one iteration of \texttt{DFW} yields the following improvement
\[
\dist(\hat\Phi_+, \Phi_\star) \leq \rho \cdot \dist(\hat\Phi, \Phi_\star) + \frac{C}{\sqrt{\text{\# tasks} \times \text{\# data per task}}},\quad \rho \in (0,1),\; C > 0.
\]
\end{theorem}
Critically, the second term of the right hand side scales jointly in the number of tasks and datapoints per task, whereas naively implementing other methods may be bottlenecked by a term that scales solely with the amount of data for a single task, which leads to suboptimal sample-efficiency.









\subsection{Related Work}

\noindent\textbf{Multi-task linear regression:} Directly related to our work are results demonstrating the benefits of multi-task learning for linear regression~\citep{maurer2016benefit, bullins2019generalize, du2020few, tripuraneni2021provable, collins2021exploiting, thekumparampil2021sample}, under the assumption of a shared but unknown linear feature representation. In particular, our proposed algorithm is adapted from the alternating optimization scheme independently in \cite{nayer2022fast} and \cite{collins2021exploiting}, and extends these results to the vector measurement setting and introduces algorithmic modifications to extend its applicability to non-iid and non-isotropic covariates. We also highlight that in the isotropic linear regression setting, \cite{thekumparampil2021sample} provide an alternating minimization scheme that results in near minimax-optimal representation learning. However, the representation update step simultaneously accesses data across tasks, which we avoid in this work due to motivating applications, e.g.\ distributed learning, that impose locality or data privacy constraints.

\noindent\textbf{Meta/multi-task RL:} There is a wealth of literature in reinforcement learning that seeks empirically to solve different tasks with shared parameters \citep{teh2017distral, hessel2018popart,singh2020scalable, deisenroth2014multitask}. In parallel, there is a body of theoretical work which studies the sample efficiency of representation learning for RL \citep{https://doi.org/10.48550/arxiv.2106.08053, https://doi.org/10.48550/arxiv.2206.05900, https://doi.org/10.48550/arxiv.1505.06279}. This line of work considers MDP settings, and thus the specific results are often stated with incompatible assumptions (such as bounded states/cost functions and discrete action spaces), and are suboptimal when instantiated in our setting.

\noindent\textbf{System identification and control:} Multi-task learning has gained recent attention in controls, e.g.\ for adaptive control over similar dynamics \citep{harrison2018control, richards2021adaptive, shi2021meta, muthirayan2022meta}, imitation learning for linear systems \citep{zhang2022multitask, guo2023imitation}, and notably linear system identification \citep{li2022data, wang2022fedsysid, xin2023learning, modi2021joint, faradonbeh2022joint, chen2023multi}. In many of these works \citep{li2022data, wang2022fedsysid, xin2023learning}, task similarity is quantified by a generic norm closeness of the dynamics matrices, and thus the benefit of multiple tasks extends only to a radius around optimality. Under the existence of a shared representation, our work provides an efficient algorithm and statistical analysis to establish convergence to per-task optimality. 



\section{Problem Formulation}\label{sec: prob formulation}


\noindent\textbf{Notation:} the Euclidean norm of a vector $x$ is denoted $\norm{x}$. The spectral and Frobenius norms of a matrix $A$ are denoted $\norm{A}$ and $\norm{A}_F$, respectively. For symmetric matrices $A,B$, $A \preccurlyeq B$ denotes $B - A$ is positive semidefinite. The largest/smallest singular and eigenvalues of a matrix $A$ are denoted $\sigma_{\max}(A)$, $\sigma_{\min}(A)$, and $\lambda_{\max}(A)$, $\lambda_{\min}(A)$, respectively. The condition number of a matrix $A$ is denoted $\kappa(A) := \sigma_{\max}(A)/\sigma_{\min}(A)$. Define the indexing shorthand $[n] := \curly{1, \dots, n}$. We use $\lesssim, \gtrsim$ to omit universal numerical factors, and $\tilde\calO(\cdot)$, $\tilde\Omega(\cdot)$ to omit polylog factors in the argument.

\paragraph{Regression Model.} Let a covariate sequence (also denoted a \textit{trajectory}) be an indexed set $\scurly{x_i}_{i \geq 1} \subset \R^{d_x}$. We denote a distribution $\prob_x$ over covariate sequences, which we assume to have bounded second moments for all $i \geq 1$, i.e.\ $\Ex\brac{x_i x_i^\top}$ is finite for all $i \geq 1$. 
Defining the filtration $\scurly{\calF_i}_{i \geq 0}$ where $\calF_i := \sigma(\scurly{x_k}_{k=1}^{i+1}, \scurly{w_k}_{k=1}^i)$ is the $\sigma$-algebra generated by the covariates up to $i+1$ and noise up to $i$, we assume that $\scurly{w_i}_{i \geq 1}$ is a $\sigma_w^2$-subgaussian martingale difference sequence (MDS):
\begin{align*}
    \Ex\brac{v^\top w_i \mid \calF_{i - 1}} = 0, \quad \Ex\brac{\exp\paren{\lambda v^\top w_i} \mid \calF_{i-1}} \leq \exp\paren{\lambda^2 \norm{v}^2 \sigma_w^2}\;\text{a.s.}\; \forall \lambda\in \R, v \in \R^{d_y}.
\end{align*}
Assuming a ground truth operator $M_\star \in \R^{d_y \times d_x}$, our observation model is given by 
$$y_i = M_\star x_i + w_i, \quad i \geq 1,$$ 
for $y_i$ the \emph{labels}, and $w_i$ the \emph{label noise}. We further define $\Sigma_{x,N} := \frac{1}{N}\sum_{i=1}^{N}\Ex[x_ix_i^\top]$. When the marginal distributions of $x_i, \; i \geq 1$ are identical, we denote $\Sigma_x \equiv \Sigma_{x, N}$.



\paragraph{Multi-Task Operator Recovery.}

We consider the following instantiation of the above linear operator regression model over multiple tasks. In particular, we consider heterogeneous data $\scurly{(\xhit, \yhit)}_{i=1, t=1}^{N, T}$, consisting of trajectories of length $N$, generated independently across $t = 1,\dots, T$ task distributions. For notational convenience, we assume that the length of trajectories $N$ is the same across training tasks. For each task $t$, the observation model is
\begin{equation}\label{eq: data generating process}
    \yhit = M^{(t)}_\star \xhit + \whit,
\end{equation}
where $M^{(t)}_\star = \Fhstar \Phi_\star$ admits a decomposition into a ground-truth representation $\Phi_\star \in \R^{r \times d_x}$ common across all tasks $t \in [T]$ and a task-specific weight matrix $\Fhstar \in \R^{d_y \times r}$, $r \leq d_x$. We denote the joint distribution over covariates and observations $\scurly{\xhit, \yhit}_{i \geq 1}$ by $\prob_{x,y}^{(t)}$. We assume that the representation $\Phi_\star$ is normalized to have orthonormal rows to prevent boundedness issues, since $\Fhstar' = \Fhstar Q^{-1}$, $\Phi_\star' = Q \Phi_\star$ are also valid decompositions for any invertible $Q \in \R^{r \times r}$. To measure closeness of an approximate representation $\hat \Phi$ to optimality, we define a subspace metric.
\begin{definition}[Subspace Distance  \citep{stewart1990matrix, collins2021exploiting}]\label{def: subspace distance}
    Let $\Phi, \Phi_\star \in \R^{r \times d_x}$ be matrices whose rows are orthonormal. Furthermore, let $\Phiperp \in \R^{(d_x - r) \times d_x}$ be  a matrix such that $\bmat{\Phi_\star^\top & \Phiperp^\top}$ is an orthogonal matrix. Define the distance between the subspaces spanned by the rows of $\Phi$ and $\Phi_\star$ by
    \begin{align}\label{eq: subspace distance}
        \dist(\Phi, \Phi_\star) &:= \snorm{\Phi \Phiperp^\top}_2
    \end{align}
\end{definition}
In particular, the subspace distance quantitatively captures the alignment between two subspaces, interpolating smoothly between  $0$ (occurring iff $\Span(\Phi_\star) = \Span(\hat\Phi)$) and $1$ (occurring iff $\Span(\Phi_\star) \perp \Span(\hat\Phi)$).
We define the task-specific stacked vector notation by capital letters, e.g.,
\[
\Xh = \bmat{x^{(t)}_1 & \cdots & x^{(t)}_i &  \cdots & x^{(t)}_N}^\top \in \R^{N \times d_x}.
\]
The goal of multi-task operator recovery is to estimate $\scurly{\Fhstar}_{t=1}^T$ and $\Phi_\star$ from data collected across multiple tasks $\scurly{(\xhit, \yhit)}_{i=1}^{N}$, $t =1,\dots, T$. Some prior works \citep{maurer2016benefit, du2020few, zhang2022multitask} assume access to an empirical risk minimization oracle, i.e.\ access to 
\[
\scurly{\hat F^{(t)}}_{t=1}^{T}, \hat\Phi \in \argmin_{\scurly{F^{(t)}}, \Phi} \sum_{t=1}^T\sum_{i=1}^N \norm{\yhit - F^{(t)} \Phi \xhit}^2,
\]
focusing on the statistical generalization properties of an ERM solution. However, the above optimization is non-convex even in the linear setting, and thus it is imperative to design and analyze efficient algorithms for recovering optimal matrices $\scurly{\Fhstar}_{t=1}^T$ and $\Phi_\star$. 
To address this problem in the linear regression setting, various works, e.g.\ \texttt{FedRep} \citep{collins2021exploiting}, \texttt{AltGD-Min} \citep{nayer2022fast}, propose an alternating minimization-descent scheme, where on a fresh data batch, the weights $\scurly{\hat F^{(t)}}$ are computed on task-specific (``local'') data via least-squares, and an estimate of the representation gradient is subsequently computed with respect to task-specific data and averaged across tasks to perform gradient descent on the representation parameters. 
This algorithmic framework is intuitive, and thus forms a reasonable starting point toward a provably sample-efficient algorithm in our setting.

\section{Sample-Efficient Linear Representation Learning}\label{sec: algorithm guarantees}

We begin by describing the vanilla alternating minimization-descent scheme proposed in \citet{collins2021exploiting}. We show that in our setting with label noise and non-isotropy, interaction terms arise in the representation gradient, which cause biases to form that do not scale down with the number of tasks $T$. In \S\ref{ss: DFW}, we propose alterations to the scheme to remove these biases, which we then show in \S\ref{ss: algorithm guarantees} lead to fast convergence rates that allow us to recover near-oracle ERM generalization bounds.

\subsection{Perils of (Vanilla) Gradient Descent on the Representation}\label{ss: perils of vanilla gradient}


We begin with a summary of the main components of an alternating minimization-descent method analogous to \texttt{FedRep}~\citep{collins2021exploiting} and \texttt{AltGD-Min}~\citep{nayer2022fast}. During each optimization round, a new data batch is sampled for each task: $\scurly{(\xhit,\yhit)}_{i=1}^{N}$, $t \in [T]$. We then compute 
task-specific weights $\hat F^{(t)}$ on the corresponding dataset, keeping the current representation estimate $\hat\Phi$ fixed. For example, $\hat F^{(t)}$ may be the least-squares weights conditioned on $\hat\Phi$ \citep{collins2021exploiting}. Define $\zhit := \hat\Phi \xhit$, and the empirical covariance matrices $\hatSigmaxhNT := \tfrac{1}{N}\Xh^\top \Xh$, $\hatSigmazhNT := \tfrac{1}{N}\Zh^\top \Zh$. The least squares solution $\hat F^{(t)}$ is given by the convex quadratic minimization
\ifshort{$\hat F^{(t)} = \argmin_{F} \sum_{i=1}^N \|{\yhit - F\zhit} \|^2$.}
{\begin{align}\label{eq: least squares weight matrix}
    \hat F^{(t)} &= \argmin_{F} \sum_{i=1}^N \norm{\yhit - F\zhit}^2 \nonumber\\
    \begin{split}
    = \Fh_\star \Phi_\star \Xh^\top \Zh (\hatSigmazhNT)^{-1} + \Wh^\top \Zh (\hatSigmazhNT)^{-1},
    \end{split}
\end{align}
where we derive \eqref{eq: least squares weight matrix} through standard matrix calculus \citep{petersen2008matrix} and expanding \eqref{eq: data generating process}.}
For each task, we then fix the weight matrix $\hat F^{(t)}$ and perform a descent step with respect to the representation conditioned on the local data. The resulting representations are averaged across tasks to form the new representation. When the descent direction is the gradient, the update rule is given by
\begin{align}
    \overline \Phi_+^{(t)} &= \hat\Phi - \frac{\eta}{2N} \nabla_{\Phi} \sum_{i=1}^N  \norm{\yhit - \hat F^{(t)}\hat\Phi \xhit}^2, \quad
    \overline \Phi_+ = \frac{1}{T}\sum_{t=1}^T \overline \Phi_+^{(t)} \label{eq: rep vanilla grad update task-conditioned}
\end{align}
where $\eta > 0$ is a given step size.  We normalize $\overline\Phi_+$ to have orthonormal rows, e.g.\ by (thin/reduced) QR decomposition \citep{tref2022numerical}, to produce the final output $\hat\Phi_+$, i.e.\ $\overline \Phi_+ = R \hat\Phi_+$, $R \in \R^{r \times r}$, leading to 
\begin{align}\label{eq: FedRep rep update} 
    R \hat\Phi_+ 
    &= \hat\Phi - \frac{\eta}{T}\sum_{t=1}^T \mathopen{}\hat F^{(t)}\mathclose{}^\top \paren{\hat F^{(t)} \hat\Phi - \Fh_\star \Phi_\star} \hatSigmaxhNT - \frac{\eta}{NT}\sum_{t=1}^T \mathopen{}\hat F^{(t)}\mathclose{}^\top \Wh^\top \Xh.
\end{align}
As in \cite{collins2021exploiting}, we right-multiply both sides of \eqref{eq: FedRep rep update} by $\Phiperp^\top$, recalling $\|\Phi \Phiperp^\top\|_2 =: \dist(\Phi, \Phi_\star)$. Crucially, \citet{collins2021exploiting} assume $\xhi$ has mean $0$ and identity covariance, and $\whit \equiv 0$ across $i,t$. 
\ifshort{Therefore, the label noise terms $\mathopen{}\hat F^{(t)}\mathclose{}^\top \Wh^\top \Xh$ disappear, and the sample covariance for each task $\hatSigmaxhNT$ concentrates to identity, such that $\Phi_\star \hatSigmaxhNT \Phiperp^\top \approx 0$. Under appropriate choice of $\eta$ and bounding the effect of the orthonormalization factor $R$, linear convergence to the optimal representation can be established.}
{Therefore, the label noise terms $\mathopen{}\hat F^{(t)}\mathclose{}^\top \Wh^\top \Xh$ disappear, and the sample covariance for each task $\hatSigmaxhNT$ concentrates to identity. Under these assumptions, we get
\begin{align*}
    \norm{R \hat\Phi_+ \Phiperp^\top} &= \norm{\hat\Phi \Phiperp^\top - \frac{\eta}{T}\sum_{t=1}^T \mathopen{}\hat F^{(t)}\mathclose{}^\top \paren{\hat F^{(t)} \hat\Phi - \Fh_\star \Phi_\star} \hatSigmaxhNT \Phiperp^\top} \\
    &\lesssim \underbrace{\norm{I - \frac{\eta}{T}\sum_{t=1}^T \mathopen{}\hat F^{(t)}\mathclose{}^\top \hat F^{(t)}}}_{\text{Contraction term}}\dist\paren{\hat\Phi,\Phi_\star} + \underbrace{\calO\paren{\frac{1}{T}\sum_{t=1}^T \norm{\hatSigmaxhNT - I_{d_x}}}}_{\text{Covariance concentration term}}.
\end{align*}
where we note $\Phi_\star \Phiperp^\top = 0$. Under appropriate choice of $\eta$ and bounding the effect of the orthonormalization factor $R$, linear convergence to the optimal representation can be established.}
However, two issues arise when label noise $\whit$ is introduced and when $\xhit$ has non-identity covariance.
\begin{enumerate}[left=0pt, noitemsep]
    \item When \emph{label noise} $\whit$ is present, since $\hatFh$ is computed on $\Yh,\Xh$, the gradient noise term is generally biased: $\frac{1}{NT}\Ex\sbrac{\hatFh \Wh^\top \Xh} \neq 0$. Even in the simple case that all task distributions $\Phxy$ are identical, $\frac{\eta}{NT}\sum_{t=1}^T \mathopen{}\hat F^{(t)}\mathclose{}^\top \Wh^\top \Xh$ concentrates to its bias, and thus for large $T$ the size of noise term is bottlenecked at $\frac{\eta}{NT}\Ex\brac{\norm{\hatFh^\top \Wh^\top \Xh}}$. This critically causes the noise term to lose scaling in the \emph{number of tasks} $T$, even when the tasks are identical.
    \item When $\xhit$ has \emph{non-identity covariance},
    the decomposition into a contraction and covariance concentration term no longer holds, since generally $\Phi_*\Ex\sbrac{\hatSigmaxhNT}\Phiperp^\top \neq 0$. This causes a term whose norm otherwise concentrates around $0$ in the isotropic case to scale with $\lambda_{\max}(\hatSigmaxhNT) - \lambda_{\min}(\hatSigmaxhNT)$ in the worst case. 
    \ifshort{}
    {Unlike prior work that assumes identical distribution of covariates $\xhit$ across tasks, this issue cannot be circumvented by whitening the covariates $\xhit$, as shifting the task-specific covariance factor to the operator $\Fh_\star \Phi_\star \mathopen{}\Sigma_{x}^{(t)}\mathclose{}^{1/2}$ in general ruins the shared representation spanned by $\Phi_\star$.} 
\end{enumerate}
This motivates modifying the representation update beyond following the vanilla stochastic gradient.

\ifshort{\vspace{-0.2cm}}{}
\subsection{A Task-Efficient Algorithm: \texttt{De-bias \& Feature-whiten}} \label{ss: DFW}

In the previous section, we identified two fundamental issues: 1.\ the bias introduced by computing the least squares weights and representation update on the same data batch, and 2.\ the nuisance term introduced by non-identity second moments of the covariates $\xhit$. Toward addressing the first issue, we introduce a ``de-biasing'' step, where each agent computes the least squares weights $\hatFh$ and the representation update on \textit{independent} batches of data, e.g.\ disjoint subsets of trajectories.
To address the second issue, we introduce a  ``feature-whitening'' adaptation \citep{lecun2002efficient}, where the gradient estimate sent by each agent is pre-conditioned by its inverse sample covariance matrix. Combining these two adaptations, the representation update becomes
\begin{align}\label{eq: update rule FW IID case}
    R \hat\Phi_+ &= \hat\Phi - \frac{\eta}{T}\sum_{t=1}^T \mathopen{}\hat F^{(t)}\mathclose{}^\top \paren{\hat F^{(t)} \hat\Phi - \Fh_\star \Phi_\star} - \frac{\eta}{T}\sum_{t=1}^T \mathopen{}\hat F^{(t)}\mathclose{}^\top \Wh^\top \Xh \paren{\hatSigmaxhNT}^{-1},
\end{align}
where we assume $\scurly{\hatFh}$ are computed on independent data using the aforementioned batching strategy. We comment that this ``pre-conditioning by the inverse sample covariance'' step bears striking resemblance to various algorithms applied to dynamical systems, e.g.\ Quasi-Newton method for Generalized Linear Models \citep{kowshik2021near}, natural policy gradient for linear-quadratic systems \citep{fazel2018global}. Curiously, the various motivations of this step differ entirely between all of these works; for example, the purpose of this step in our work arises even for independent data. When $\xhit, \whit$, $t = 1,\dots, T$, are all mutually independent, then the first two terms of the update form the contraction, and the last term is an average of \emph{zero-mean} least-squares-error-like terms over tasks, which can be studied using standard tools \citep{abbasi2011online, abbasi2013online}. This culminates in convergence rates that scale favorably with the number of tasks (\S\ref{ss: algorithm guarantees}).
To operationalize our proposed adaptations, let $D^{(t)} = \scurly{(\xhit, \yhit)}_{ i=1}^{N}$, $t\in [T]$, be a dataset available to each agent. For the weights de-biasing step, we sub-sample trajectories $\calN_1 \subset [N],\;\abs{\calN_1} := N_1$. For each agent, we compute least-squares weights from $\calN_1$. We then sub-sample trajectories $\calN_2 \subset  [N]\setminus \calN_1,\;\abs{\calN_2} = N_2$, and compute the task-conditioned representation gradients from $\calN_2$.
\begin{align}\label{eq: debiased least squares}
    \hat F^{(t)} = \argmin_{F} \sum_{i\in \calN_1} \norm{\yhit - F\zhit}^2, \quad \hat\calG_{\calN_2}^{(t)} =  \nabla_{\Phi} \frac{1}{2} \sum_{i\in \calN_2} \norm{\yhit - \hat F^{(t)}\hat\Phi \xhit}^2.
\end{align}
Lastly, each agent updates its local representation via a feature-whitened gradient step to yield $\bar\Phi_+^{(t)}$. The global representation update is computed by averaging the updated task-conditioned representations $\bar\Phi_+^{(t)}$ and performing orthonormalization:
\begin{align}
    \bar\Phi_+^{(t)} := \hat\Phi - \eta \hat\calG_{\calN_2}^{(t)} \paren{\hat\Sigma^{(t)}_{x, \calN_2}}^{-1},\quad R\hat\Phi_+ &= \frac{1}{T} \sum_{t=1}^T \bar\Phi_+^{(t)} \label{eq: representation update IID setting} \text{ s.t. }\hat\Phi_+^\top \hat\Phi_+ = I_r\\
    \iff \hat\Phi_+, R &= \texttt{thin\_QR}\paren{\frac{1}{T} \sum_{t=1}^T \bar\Phi_+^{(t)}}, \nonumber
\end{align}
We summarize the full algorithm in \Cref{alg: multi-task alt min descent}. The above de-biasing and feature whitening steps ensure that the expectation of the representation update \eqref{eq: update rule FW IID case} is a contraction (with high probability):
\begin{align}
    \begin{split} \label{eq: contraction}
        &R \hat\Phi_+ \Phiperp^\top  =\paren{I - \frac{\eta}{T}\sum_{t=1}^T \hatFh^\top \hatFh} \hat\Phi\Phiperp^\top - \frac{\eta}{T}\sum_{t=1}^T \mathopen{}\hat F^{(t)}\mathclose{}^\top \Wh^\top \Xh \paren{\hatSigmaxhNT}^{-1}\Phiperp^\top \\
        \implies \;&\Ex\brac{\dist(\hat\Phi_+, \Phi_\star)} = \Ex\brac{\norm{R^{-1}\paren{I - \frac{\eta}{T}\sum_{t=1}^T \hatFh^\top \hatFh}}} \dist(\hat\Phi, \Phi_\star),
    \end{split}
\end{align}
where the task and trajectory-wise independence ensures that the variance of the gradient scales inversely in $NT$.

\begin{algorithm}[t]
    \caption{\texttt{De-biased \& Feature-whitened} (DFW) Alt.\ Minimization-Descent}
    \label{alg: multi-task alt min descent}
    \begin{algorithmic}[1]
        \State \textbf{Input:} step sizes $\curly{\eta_k}_{k \geq 1}$, batch sizes $\curly{N_k}_{k \geq 1}$, initial estimate $\hat\Phi_0$.
        \For {$k = 1, \dots, K$}
            \For {$t \in [T]$ \textbf{(in parallel)}}
                \State Obtain samples $\scurly{(\xhit, \yhit)}_{i=1}^{N_k}$.
                \State Partition trajectories $[N_k] = \calN_{k,1} \sqcup \calN_{k,2}$.
                \State Compute $\hat F^{(t)}_k$, e.g.\ via least squares on $\calN_{k,1}$ \eqref{eq: debiased least squares}.
                \State Compute task-conditioned representation gradient $\hat\calG_{\calN_{k,2}}^{(t)}$ on $\calN_{k,2}$ \eqref{eq: debiased least squares}.
                \State Compute task-conditioned representation update $\bar\Phi^{(t)}_k$ \eqref{eq: representation update IID setting}.
            \EndFor
            \State $\hat\Phi_k, \_ \leftarrow \texttt{thin\_QR}\paren{\frac{1}{T} \sum_{t=1}^{T} \bar\Phi_{k}^{(t)}}$.
        \EndFor
        \State \Return Representation estimate $\hat\Phi_K$.
    \end{algorithmic}
\end{algorithm}

\ifshort{}
{
\begin{remark}[Choice of weights $\hatFh$ vs.\ descent rate]\label{remark: least square weights}
    By observing the contraction expression \eqref{eq: contraction}, the contraction rate is seemingly solely controlled by the (average) conditioning of the weight matrices $\hatFh$. Since the choice of algorithm for computing $\hatFh$ is user-determined, this motivates choosing well-conditioned $\hatFh$. However, the hidden trade-off lies in the orthonormalization factor $R$; arbitrary $\hatFh$ may lead to $R$ that undoes progress. As in \cite{collins2021exploiting}, we analyze $\hatFh$ generated by representation-conditioned least squares \eqref{eq: debiased least squares}, but an optimal balance between conditioning of $\hatFh$ and $R$ can be struck by $\ell^2$-regularized least squares weights $\hatFh(\lambda)$ (see, e.g.\ \cite{hsu2012random}).
\end{remark}
}

\subsection{Algorithm Guarantees}\label{ss: algorithm guarantees}

We present our main result in the form of convergence guarantees for \Cref{alg: multi-task alt min descent}. 
We begin by defining a standard measure of dependency along covariate sequences via $\beta$-mixing.
\begin{definition}[$\beta$-mixing]\label{assumption: beta-mixing}
Let $\scurly{x_i}_{i \geq 1}$ be a $\R^d$-valued discrete-time stochastic process adapted to filtration $\scurly{\calF_i}_{i = 1}^\infty$. 
We denote the stationary distribution $\nu_\infty$. We define the $\beta$-mixing coefficient
\begin{equation}
    \beta(k) := \sup_{i \geq 1} \Ex_{\curly{x_\ell}_{\ell=1}^i}\brac{\norm{\prob_{x_{i+k}}(\cdot \mid \calF_i) - \nu_\infty }_{\mathrm{tv}}},
\end{equation}
where $\norm{\cdot}_{\mathrm{tv}}$ denotes the total variation distance between probability measures.
\end{definition}
Intuitively, the $\beta$-mixing coefficient measures how quickly on average a process converges to its stationary distribution along any sample path.
To instantiate our bounds, we make the following assumptions on the covariates.

\begin{assumption}[Subgaussian covariates, geometric mixing]\label{assumption: subgaussian covariate}
Given number of tasks $T$ and per-task samples $N$, we assume the marginal distributions of $\xhit$ to be identical with zero-mean, covariance $\Sigmaxh$, and to $\gamma^2$-subgaussian across all $i \in [N], t \in [T]$:
\[
\Ex[\xhit]=0, \quad \Ex\brac{\exp\paren{\lambda v^\top \xhit}} \leq \exp\paren{\lambda^2 \norm{v}^2 \gamma^2}\quad \text{a.s. } \forall \lambda \in \R, v \in \R^{d_x}.
\]
Furthermore, we assume the process $\scurly{\xhit}_{i\geq 1}$ is geometrically $\beta$-mixing with $\betah(k) \leq \Gammah \muh^k$, $\Gammah > 0$, $\muh \in [0, 1)$, for each task $t \in [T]$. Lastly, we define $\tau_{\mathsf{mix}}^{(t)} := \paren{\frac{\log(\Gammah N/\delta)}{\log(1/\muh)} \vee 1}$.
\end{assumption}
Notably, these assumptions subsume fundamental problems in learning over (stable) dynamical systems, in particular linear system identification and imitation learning, where non-iid and non-isotropy across tasks are unavoidable. We discuss these instantiations in depth in \Cref{appx: linear systems case-study}. As in prior work, our final convergence rates depend on a notion of task diversity.
\begin{definition}[Task diversity]
We define the quantities
\begin{align}\label{eq: task diversity constants}
    \lambda^{\bfF}_{\min} := \lambda_{\min} \paren{\frac{1}{T} \sum_{t=1}^T\Fhstar^\top \Fhstar}, \quad \lambda^\bfF_{\max} := \lambda_{\max} \paren{\frac{1}{T} \sum_{t=1}^T\Fhstar^\top \Fhstar}.
\end{align}
\end{definition}

As hinted by \Cref{eq: contraction}, our proof strategy boils down to bounding the various terms in the update rule of \texttt{DFW}; in particular, the ``contraction factor'' $I - \frac{\eta}{T}\sumT \hatFh^\top \hatFh$, the task-averaged noise term $\frac{1}{T} \sumT \hatFh^\top \Wh^\top \Xh (\hatSigmaxhNT)^{-1}$, and lastly the effect of orthnormalization $R$. Starting with the contraction factor, we observe expanding the least-squares weights yields:
\begin{align*}
    \hatFh &= \Fh_\star \Phi_\star \Xh^\top \Zh (\hatSigmazhNT)^{-1} + \Wh^\top \Zh (\hatSigmazhNT)^{-1} \\
    &= \Fhstar \Phi_\star \hat\Phi^\top + \Fhstar \Phi_\star \hatPhiperp^\top \hatPhiperp \Xh^\top \Zh (\hatSigmazhNT)^{-1} + \Wh^\top \Zh (\hatSigmazhNT)^{-1}.
\end{align*}
Intuitively speaking, the least-squares weights decomposes into a term proportional to $\Fhstar$, an error term arising from misspecification scaling with $\snorm{\Phi_\star \hatPhiperp^\top} = \dist(\hat\Phi, \Phi_\star)$, and a zero-mean least-squares error term scaling with $\sigmahw$. Therefore, inverting bounds on the error terms into burn-in conditions, we get after some computation the following bound.
\begin{lemma}[Contraction factor bound]\label{lem: contraction factor bound main paper}
    Let \Cref{assumption: subgaussian covariate} hold. If the following burn-in conditions hold
    \begin{align*}
        \dist(\hat\Phi, \Phi_\star) &\leq \frac{1}{100} \sqrt{\frac{\bfFmin}{\bfFmax}} \max_t \kappa\mem\paren{\Sigmaxh}^{-1} \\
        N_1 &\gtrsim \max_t \tau_{\mathsf{mix}}^{(t)} \cdot \max\curly{\gamma^4\paren{r + \log(T/\delta)},  \bfFmin^{-1}\frac{1}{T}\sumT \frac{\sigmahw^2 (d_y + r + \log(T/\delta))}{\lambda_{\min}(\Sigmaxh) } },
    \end{align*}
    then, for step-size satisfying $\eta \leq 0.956 \bfFmax^{-1}$, with probability at least $1 - \delta$, we have
    \begin{align*}
        \norm{I_{d_x} - \eta \frac{1}{T}\sum_{t=1}^T\hatFh^\top \hatFh} &\leq \paren{1 -  0.954 \eta \bfFmin }.
    \end{align*}
\end{lemma}
Setting $\eta \approx \bfFmax^{-1}$, we observe that the bound on the contraction factor is approximately $1 - c \frac{\bfFmin}{\bfFmax}$, which is the best one can hope for in the spectral norm. Furthermore, we note that past the burn-in, the contraction rate is independent of the data-size $N_1 = \abs{\calN_1}$ used to compute the weights, implying that contraction holds as long as the least-squares weights are ``good enough'' (tying back to \Cref{remark: least square weights}), further implying $N_1$ can be held fixed across rounds of \texttt{DFW}.

To bound the \texttt{DFW} noise term, we observe that for each task, $\hatFh^\top \Wh^\top \Xh (\hatSigmaxhNT)^{-1}$ is an $r \times d_x$ matrix-valued self-normalized martingale \citep{abbasi2013online}. Importantly, by the de-biasing step, the weights $\hatFh$ are mutually independent of of the processes $\Wh, \Xh$ in the independent covariates setting. Similarly to the contraction factor, assuming a burn-in on $\dist(\hat\Phi, \Phi_\star)$ and $N_1$, the subgaussian constant of $\hatFh^\top \whit$ can be bounded by, say, $2 \snorm{\Fhstar}_2^2 \sigmahw^2$. By realizability \eqref{eq: data generating process}, we see that in the independent covariates setting, $\hatFh^\top \Wh^\top \Xh (\hatSigmaxhNT)^{-1}$ is a \textit{zero-mean} process that is bounded with high-probability. Therefore, applying a matrix Hoeffding inequality \citep{tropp2011user} across tasks $T$ crucially yields a bound on the noise term that benefits from more tasks $T$.
\begin{proposition}[Noise term bound]\label{prop: DFW update noise term final bound main paper}
    Let \Cref{assumption: subgaussian covariate} hold. Assume
    \begin{align*}
        \dist(\hat\Phi, \Phi_\star) &\leq \max_t \frac{1}{100} \kappa\mem\paren{\Sigmaxh}^{-1} \\
        N_1 &\gtrsim \max_t \tau_{\mathsf{mix}}^{(t)} \cdot \max\curly{\gamma^4 \paren{r + \log(T/\delta)}, \max_t \frac{\sigmahw^2}{\snorm{\Fhstar}^2\lambda_{\min}(\Sigmaxh)} \paren{d_y + r + \log(T/\delta)}}, \\
        N_2 &\gtrsim \max_t \tau_{\mathsf{mix}}^{(t)} \cdot \gamma^4 \paren{d_x + \log(T/\delta)}.
    \end{align*}
    Then, with probability at least $1 - \delta$,
    \begin{align*}
        \norm{\frac{1}{T}\sum_{t=1}^T \mathopen{}\hat F^{(t)}\mathclose{}^\top \Whtwo^\top \Xhtwo \paren{\hat\Sigma^{(t)}_{x, \calN_2}}^{-1}} &\lesssim
        \sigmaavg \sqrt{\frac{d_x + \log(T/\delta)}{TN_2} \log\paren{\frac{d_x}{\delta}}},
    \end{align*}
    where $\sigmaavg := \sqrt{\frac{1}{T}\sumT \frac{\tau_{\mathsf{mix}}^{(t)} 
 \sigmahw^2 \snorm{\Fhstar}^2}{\lambda_{\min}(\Sigmaxh)}}$ is the \emph{task-averaged} noise-level.
\end{proposition}
The final piece of the proof lies in bounding the orthonormalization factor $R$. A key observation is that $RR^\top = (R\hat\Phi_+)(R\hat\Phi_+)^\top$, where $R\hat\Phi_+$ is precisely the output of the preconditioned gradient step on $\hat\Phi$ \eqref{eq: update rule FW IID case}. Roughly speaking, we observe that the RHS of \eqref{eq: update rule FW IID case} is composed of three terms, the first of which $\hat\Phi$ satisfies $\hat\Phi\hat\Phi^\top = I_r$, the second which scales with $\frac{1}{T}\sumT \hatFh\hat\Phi - \Fhstar\Phi_\star$, and the third that is the task-averaged self-normalized martingale term. Therefore, by re-invoking tools used in \Cref{lem: contraction factor bound main paper} and \Cref{prop: DFW update noise term final bound main paper}, we find that under similar burn-in conditions, the orthogonalization factor is a small perturbation to identity, thus leaving the contraction rate and noise term essentially unaffected.

\begin{lemma}\label{lem: orthogonalization factor bound main paper}
    Let \Cref{assumption: subgaussian covariate} hold. Let the following burn-in conditions hold:
    \begin{align*}
        \dist(\hat\Phi, \Phi_\star) &\leq \frac{1}{100} \sqrt{\frac{\bfFmin}{\bfFmax}} \max_t \kappa\mem\paren{\Sigmaxh}^{-1} \\
        N_1 &\gtrsim \max_t \tau_{\mathsf{mix}}^{(t)} \cdot \max \curly{ \gamma^4 \paren{r + \log(T/\delta)}, \;\; \overline{\sigma}_{\bfF}^2\paren{d_y + r + \log(T/\delta)}} , \\
        N_2 &\gtrsim \max_t \tau_{\mathsf{mix}}^{(t)} \cdot \max\curly{\gamma^4 \paren{d_x + \log(T/\delta)},\;\;  \bfFmin^{-1} \frac{\sigmaavg^2}{T} (d_x + \log(T/\delta)) \log\paren{\frac{d_x}{\delta}}  },
    \end{align*}
    where $\overline{\sigma}_{\bfF}^2 := \max\Big\{\max_t \frac{\sigmahw^2}{\snorm{\Fhstar}^2\lambda_{\min}(\Sigmaxh)},  \frac{1}{T}\sumT \frac{\sigmahw^2}{\bfFmin\lambda_{\min}(\Sigmaxh) }\Big\}$.
    Then, given $\eta \leq 0.956 \bfFmax^{-1}$, with probability at least $1 - \delta$, we have the following bound on the orthogonalization factor $R$:
    \begin{align*}
        \norm{R^{-1}} &\leq \paren{1 - 0.0575 \eta \bfFmin}^{-1/2}.
    \end{align*}
\end{lemma}
Putting all the pieces together, we now present our main result regarding the subspace distance improvement from running one iteration of \texttt{DFW}.
\begin{theorem}[Main result]\label{thm: descent guarantee iid}
    Let \Cref{assumption: subgaussian covariate} hold, and $\sigmaavg^2, \overline{\sigma}_{\bfF}^2$ be as defined in \Cref{prop: DFW update noise term final bound main paper} and \Cref{lem: orthogonalization factor bound main paper}. Let the following burn-in conditions hold:
    \begin{align*}
        \dist(\hat\Phi, \Phi_\star) &\leq \frac{1}{100} \sqrt{\frac{\bfFmin}{\bfFmax}} \max_t \kappa\mem\paren{\Sigmaxh}^{-1} \\
        N_1 &\gtrsim \max_t \tau_{\mathsf{mix}}^{(t)} \cdot  \max \curly{ \gamma^4 \paren{r + \log(T/\delta)}, \;\; \overline{\sigma}_{\bfF}^2\paren{d_y + r + \log(T/\delta)}} , \\
        N_2 &\gtrsim \max_t \tau_{\mathsf{mix}}^{(t)} \cdot \max\curly{\gamma^4 \paren{d_x + \log(T/\delta)},\;\;  \bfFmin^{-1} \blue{\frac{\sigmaavg^2}{T}} (d_x + \log(T/\delta)) \log\paren{\frac{d_x}{\delta}}  },
    \end{align*}
    Then, given step-size satisfying $\eta \leq 0.956 \bfFmax^{-1}$, running an iteration of DFW yields an updated representation $\hat\Phi_+$ that satisfies with probability at least $1-\delta$:
    \begin{align*}
        \dist(\hat\Phi_+, \Phi_\star) &\leq \paren{1 - 0.897 \eta\bfFmin} \dist(\hat\Phi, \Phi_\star) + C \cdot \sigmaavg \sqrt{\frac{d_x + \log(T/\delta)}{\blue{TN_2}} \log\paren{\frac{d_x}{\delta}}},
    \end{align*}
    where $C > 0$ is a universal constant.
\end{theorem}
The full proofs of \Cref{thm: descent guarantee iid} and the component results are mapped out in \Cref{sec: full theoretical analysis}. Some comments are in order. We specifically highlight the benefit of multi-task data in blue. Furthermore, the dependence on $N_1$ appears only in the burn-in, implying the amount of data used to compute the weights $\hatFh$ need not grow, and only $N_2$--the amount of data used to perform the preconditioned gradient step--needs to grow to monotonically decrease the subspace distance. This is perhaps surprising, as this implies that reconstructing the operator from the intermediate weights and representations $\hat M^{(t)} = \hatFh \hat\Phi$ need not converge to $M^{(t)}_\star$ for \texttt{DFW} to converge to the optimal representation (cf.\ \Cref{appx: ERM transfer learning}).

\begin{remark}[Initialization]
    We note that Theorem \ref{thm: descent guarantee iid} relies on the representation $\hat\Phi$ being sufficiently close to $\Phi_\star$. We do not address this issue in this paper, and refer to \citet{collins2021exploiting, tripuraneni2021provable, thekumparampil2021sample} for initialization schemes in the iid linear regression setting. Our experiments suggest that initialization is often unnecessary, which mirrors the experimental findings in \citet[Sec.\ 6]{thekumparampil2021sample}. We leave constructing an initialization scheme for our general setting, or showing whether it is unnecessary, to future work.
\end{remark}

\begin{remark}[Burn-in dependence on $d_x$]
    We remark that the burn-in requirement on $N_2$ scales linearly with the covariate dimension $d_x$. This is larger than the $\Omega(r)$ requirements in the algorithm/analysis in \cite{collins2021exploiting, thekumparampil2021sample}, which we emphasize critically \emph{impose isotropy across all covariates}. In the case of \texttt{DFW}, the $d_x$ dependence arises for a fundamental reason: when $N_2 < d_x$, then postmultiplying each task's representation gradient by $(\hatSigmaxhNT)^{\dagger}$ yields subspace distance error terms scaling as $\hat\Phi \calP_{\mathrm{range}(\hatSigmaxhNT)} \Phiperp^\top$ instead of $\hat\Phi \Phiperp^\top$. Crucially, $\hat\Phi \calP_{\mathrm{range}(\hatSigmaxhNT)} \Phiperp^\top$ is an unbiased estimator of $\hat\Phi \Phiperp^\top$ only if $\xhit$ is \emph{isotropic}, and thus improvement in the subspace distance can only be guaranteed for non-isotropic $\xhit$ when $N_2 \geq d_x$. We leave as an open question whether an efficient algorithm for linear representation learning can be posed/analyzed for non-isotropic covariates in the low per-task data regime $N < d_x$.
\end{remark}


A key benefit of having the variance term in Theorem~\ref{thm: descent guarantee iid} scale properly in $N,T$ is that we may construct representations on fixed datasets whose error scales on the same order as the oracle empirical risk minimizer by running \Cref{alg: multi-task alt min descent} on appropriately partitioned subsets of a given dataset.
\begin{corollary}[Approximate ERM]\label{cor: approximate ERM}
Let the assumptions of Theorem~\ref{thm: descent guarantee iid} hold. Let $\hat\Phi_0$ be an initial representation satisfying $\dist(\hat\Phi_0, \Phi_\star) < \nu$, and define $\rho := 1 - 0.857 \frac{\lambda_{\min}^{\bfF}}{\lambda_{\max}^{\bfF}}$. Let $\bD := \scurly{\scurly{(\xhit, \yhit)}_{i=1}^{ N}}_{t \in [T]}$ be a given dataset. There exists a partition of $\bD$ into independent batches $\bB_1, \dots, \bB_K$, such that iterating \texttt{DFW} on $\bB_k$, $k \in [K]$ yields with probability greater than $1 - \delta$:
    \begin{align}\label{eq: approx ERM}
        \dist(\hat\Phi_K, \Phi_\star)^2 \leq \tilde\calO\paren{C(\rho) \sigmaavg^2 \frac{(d_x r + \log(1/\delta))}{NT}},
    \end{align}
    where $C(\rho) > 0$ is a constant depending on the contraction rate $\rho$.
\end{corollary}
We note that the RHS of \eqref{eq: approx ERM} has the ``correct'' scaling: noise level $\sigmahw^2$ multiplied by \#~parameters of the representation, divided by the \emph{total} amount of data $NT$. 
In particular, given a fine-tuning dataset of size $N'$ sampled from a task $T+1$ that shares the representation $\Phi_\star$, computing the least squares weights $\hat F^{(T+1)}$ conditioned on $\hat\Phi_K$ yields a high-probability bound (cf.\ \Cref{appx: ERM transfer learning}) on the parameter error
\begin{align*}
    \norm{\hat F^{(T+1)} \hat\Phi_K - M_\star^{(T+1)}}_F^2 &\lesssim \dist(\hat\Phi_K, \Phi_\star)^2 + {\sigma_w^{(T+1)}}^2 \frac{d_y r + \log(1/\delta)}{N'} \\
    &\lesssim C(\rho) \frac{\sigmaavg^2 (d_x r + \log(1/\delta))}{NT} +  \frac{\mathopen{}\sigma_w^{(T+1)}\mathclose{}^2(d_y r + \log(1/\delta))}{N'},
\end{align*}
where we omit task-related quantities for clarity. We note that the above parameter recovery bound mirrors that of ERM estimates \citep{du2020few, zhang2022multitask}, where we note the latter fine-tuning term scales with $d_y r$  (the number of parameters in $F^{(T+1)}$) as opposed to $r$ in the linear regression setting ($d_y = 1$).

\ifshort{\vspace{-0.3cm}}{}
\section{Numerical Validation}\label{sec: numerical}

\ifshort{\vspace{-0.3cm}}{}








We present numerical experiments
to demonstrate the key importance of aspects of our proposed algorithm. We consider two scenarios: 1.\ linear regression with non-isotropic 
\textit{iid} data, and 2.\ linear system identification.
The linear regression experiments highlight the breakdown of standard approaches when approached with mildly non-isotropic data, thus highlighting the necessity of our proposed \texttt{De-biasing \& Feature-whitening} steps, and our system identification experiments demonstrate the applicability of our algorithm to sequentially dependent (non-isotropic) data. Additional experiments and experiment details can be found in \Cref{appx: additional experiments}.

\ifshort{\vspace{-0.3cm}}{}
\subsection{Linear Regression with IID and Non-isotropic Data}

We consider the observation model from \eqref{eq: data generating process}, where we set the operator dimensions and rank as $d_x=d_y=50$ and $r=7$.
We generate the $T$ operators using the following steps: 1.\ a ground truth representation $\Phi_\star \in \R^{r \times d_x}$ is randomly generated through applying \texttt{thin\_svd} to a random matrix with values $\iidsim \calN(0,1)$, 2.\ a nominal task weight matrix $F_0 \in \R^{d_y \times r}$ is generated with elements $\iidsim \calN(0,1)$, 3.\ task-specific weights $\Fhstar \in \R^{d_y \times r}$, $t \in [T]$ are generated by applying random rotations to $F_0$. A non-isotropic covariance matrix $\Sigma_x$ shared across all tasks is generated as 
$\Sigma_x = \frac{d_x\tilde{\Sigma}_x}{\text{Tr}(\tilde{\Sigma}_x)}$, where $\tilde{\Sigma}_x=\frac{1}{2}(U+U^{\top})$, $U=5\cdot  I_{d_x} + V$, with $V_{i,j}\iidsim\mathrm{Unif}(0,1)$. We note that by design of $U$, $\Sigma_x$ is only mildly non-isotropic.
Figure \ref{fig:linear_regression} compares the performance of Algorithm \ref{alg: multi-task alt min descent}, for a single-task ($T=1$) and multiple tasks ($T=25$), as well as standard alternating minimization-descent like \texttt{FedRep} \citep{collins2021exploiting} with $T=25$. For each optimization iteration, we sample $N=100$ fresh data per task. The figure highlights that \texttt{DFW} is able to make use of all source data to decrease variance and learn a representation to near-optimality. 
As predicted in \S\ref{ss: perils of vanilla gradient}, Figure \ref{fig:linear_regression} also shows that vanilla alternating minimization-descent is not able to improve beyond a highly suboptimal bias, despite all tasks sharing the same, rather mildly non-isotropic covariate distribution.

For a second experiment, we consider instead of applying vanilla alternating minimization-descent on a linear representation, we parameterize the representation by a neural network. In particular, we consider a ReLU-activated network with one hidden layer of dimension $64$. We keep the same data-generation, with $N=100$ fresh samples per optimization iteration per task, and compare \texttt{DFW} (on a linear representation) to AMD on the neural net representation. Since the subspace distance is no longer defined for the neural net representation, we measure optimality of the learned representations by computing the average least-squares loss with respect to a validation set generated through the nominal weight matrix $F_0$. We plot the ``optimal'' loss as that attained by the ground truth operator $F_0 \Phi_\star$. We see in \Cref{fig:NN} that, despite the much greater feature representation power of the neural net, alternating minimization-descent plateaus for many iterations just like the linear case before finding a non-linear parameter representation that allows descent to optimality, taking almost two orders of magnitude more iterations than \texttt{DFW} to reach optimality. This feature learning phase is only exacerbated when there are fewer tasks. We note that in such a streaming data setting, the high iteration complexity of AMD translates to a greatly hampered sample-efficiency compared to \texttt{DFW}.

\ifshort{\vspace{-0.3cm}}{}
\subsection{Linear System Identification}\label{ss:sysID}
\ifshort{\vspace{-0.2cm}}{}
We consider a discrete-time linear system identification (sysID) problem, with dynamics 
\[
x_{i+1} = Ax_i + Bu_i + w_i, \;\ i= 0,\dots, N-1,
\]
where $x_i$ is the state of the system and $u_i$ is the control input.
In contrast to the previous example, the covariates are now additionally non-iid due to correlation over time. In particular, we can instantiate multi-task linear sysID in the form of \eqref{eq: data generating process},
\begin{align*}
    x^{(t)}_{i+1} = M^{(t)}_{\star}z^{(t)}_i + w^{(t)}_i, \;\ i=0,\ldots, N-1
\end{align*}
where  $M^{(t)}_{\star} := [A^{(t)} \;\ B^{(t)}] = F^{(t)}\Phi_\star  \in \mathbb{R}^{d_x \times d_z}$.
The state-action pair at time instant $i$ for all tasks $t \in [T]$ is embedded as $z^{(t)}_i = [\xhit^\top \;\ \mathopen{}u^{(t)}_i \mathclose{}^\top]^\top$. The process noise $w^{(t)}_i$ and control action $u^{(t)}_i$ are assumed to be drawn from Gaussian distributions $\mathcal{N}(0,\Sigma_w)$ and $\mathcal{N}(0,\sigma^2_uI_{d_u})$, respectively, where $d_u$ represents the dimension of the control action.
We set the state dimension $d_x = 25$, control dimension $d_u = 2$, latent dimension $r=6$, horizon $N=100$, and input variance $\sigma^2_u=1$. The generation process of the ground truth system matrices $M^{(t)}_\star$ follows a similar approach as described in the linear regression problem, with the addition of a normalization step of the nominal weight matrix $F_0$ to ensure system stability for all tasks $t \in [T]$.  Furthermore, the process noise covariance $\Sigma_w$ is parameterized in a similar manner as in the linear regression example, with $U=5\cdot I_{d_x} + 2\cdot V$. 
The initial state $x^{(t)}_0$ is drawn iid across tasks from the system's stationary distribution $\mathcal{N}(0,\Sigma^{(t)}_x)$, which is determined by the solution to the discrete Lyapunov equation $\Sigma^{(t)}_x = A^{(t)}\Sigma^{(t)}_x(A^{(t)})^\top + \sigma^2_u B^{(t)}(B^{(t)})^\top + \Sigma_w$. We note this implies the covariates $\xhit$ are inherently non-isotropic and non-identically distributed across tasks. 
Figure \ref{fig:sysID} again demonstrates the advantage of leveraging multi-task data to reduce the error in computing a shared representation across the system matrices $M^{(t)}_\star$. In line with our theoretical findings, \texttt{DFW} continues to benefit from multiple tasks, even when the data is sequentially dependent. We see that \texttt{FedRep} remains suboptimal in this non-iid, non-isotropic setting.

\begin{figure}[t]
     \centering
     \begin{subfigure}[b]{0.31\textwidth}
         \centering
         \includegraphics[width=1\textwidth]{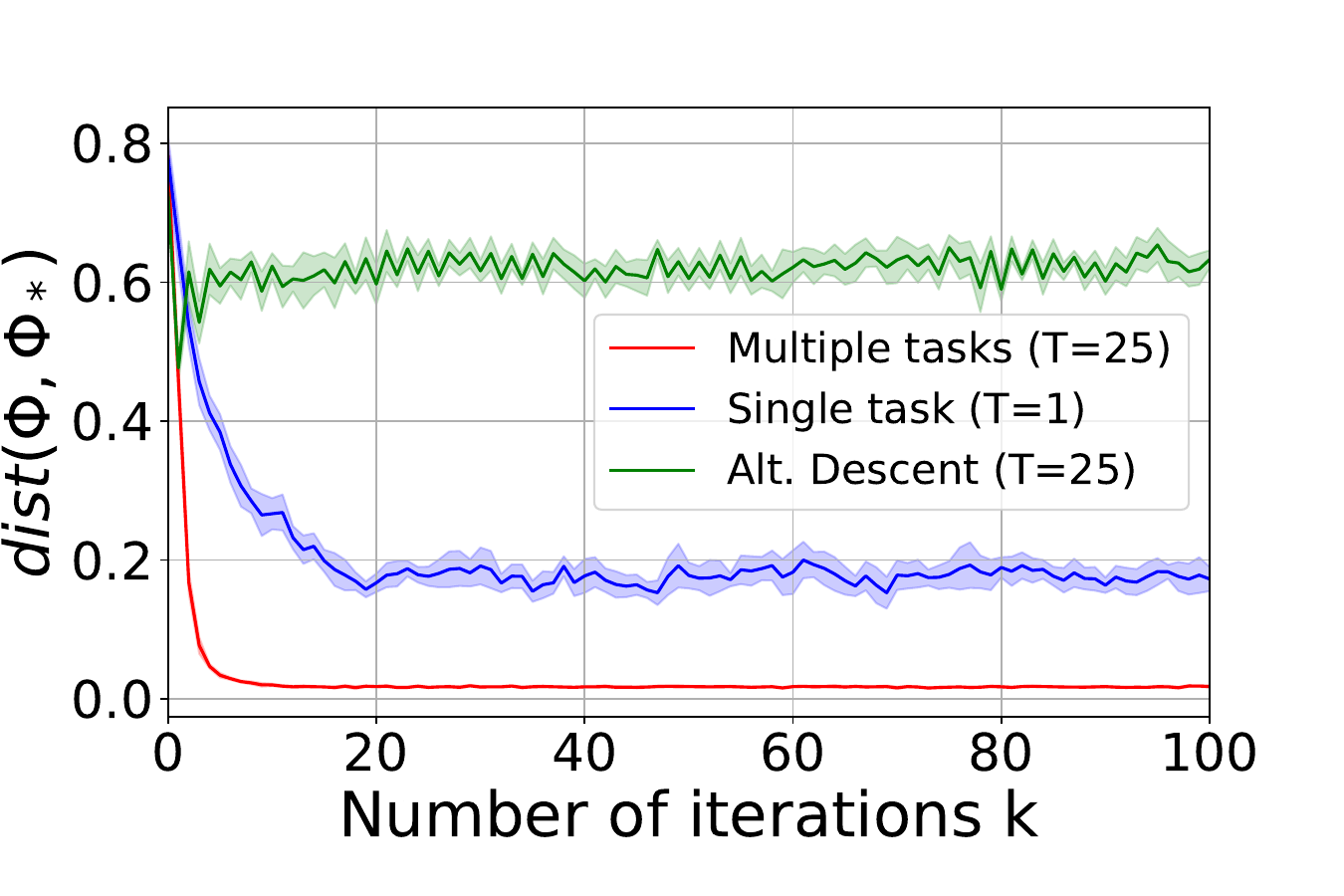}
         \caption{IID linear regression }
         \label{fig:linear_regression}
     \end{subfigure}
     \begin{subfigure}[c]{0.31\textwidth}
          \vspace{-3.2cm}
         \centering
         \includegraphics[width=1\textwidth]{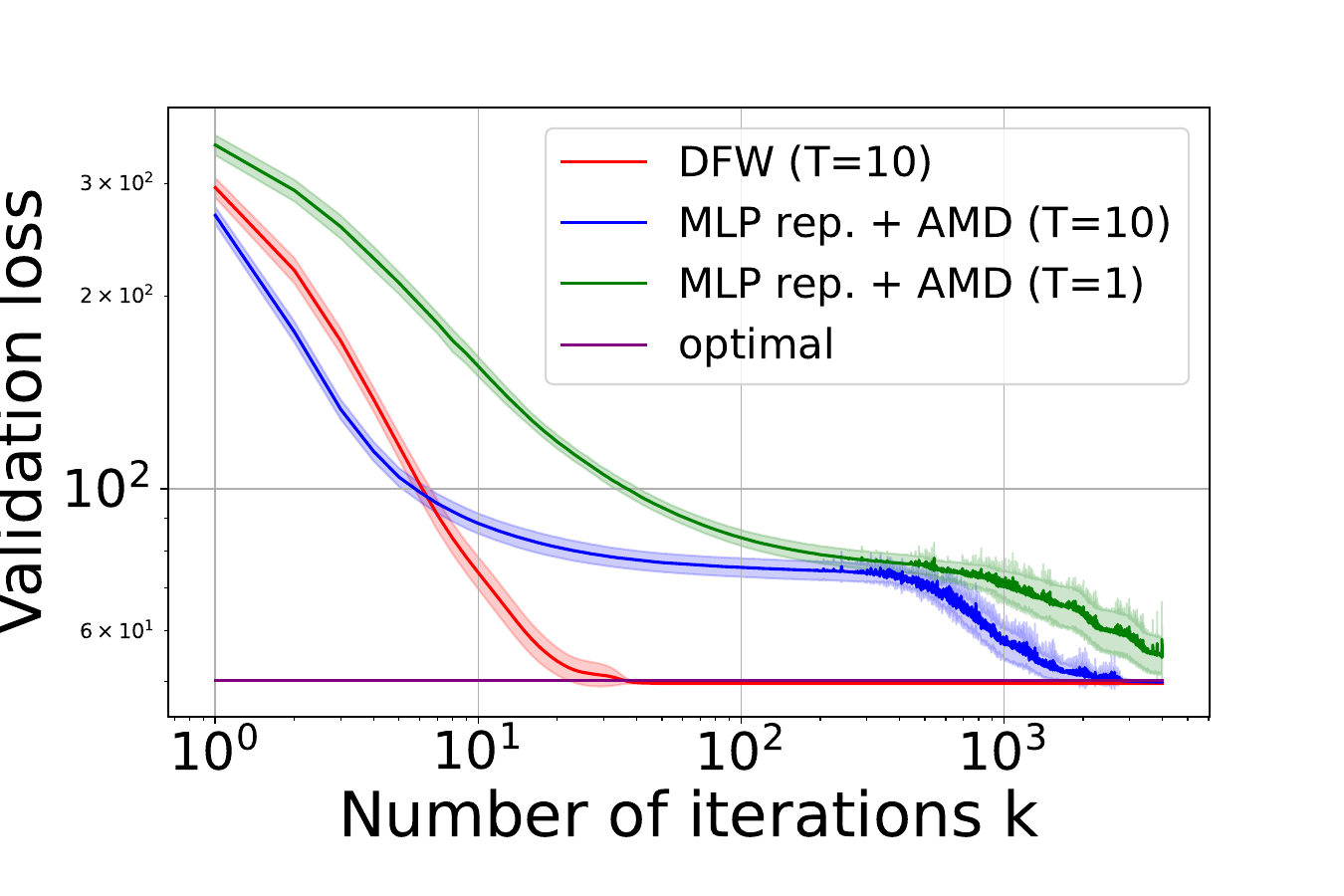}
         \caption{Neural net rep.\ + AMD}
         \label{fig:NN}
     \end{subfigure}
     \begin{subfigure}[b]{0.31\textwidth}
         \centering
         \includegraphics[width=1\textwidth]{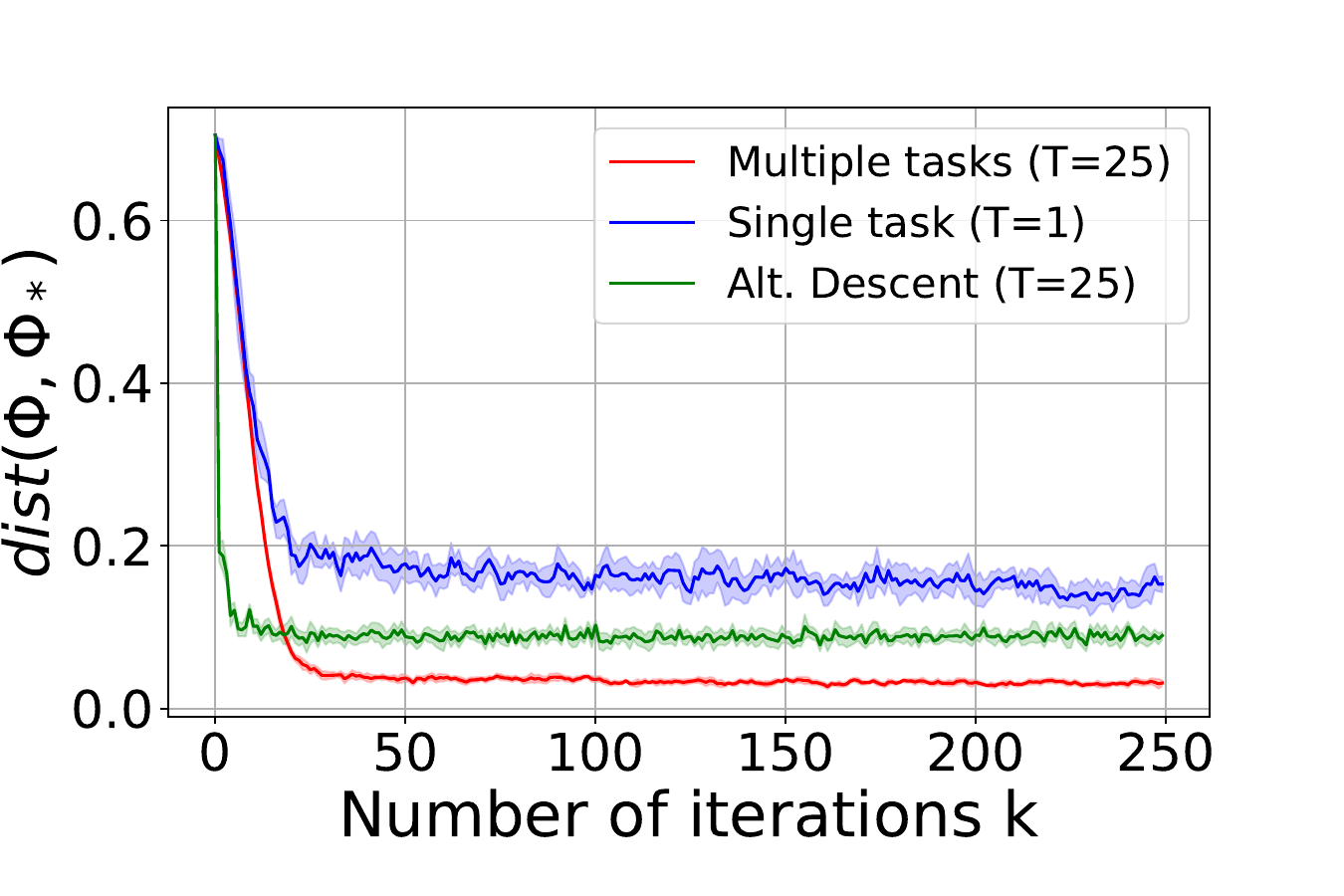}
         \caption{Linear system identification}
         \label{fig:sysID}
     \end{subfigure}
        \caption{We plot the suboptimality the current and ground truth representation with respect to the number of iterations, comparing between the single and multiple-task settings of Algorithm \ref{alg: multi-task alt min descent} and the multi-task alternating minimization-descent. We observe performance improvement and variance reduction for multi-task \texttt{DFW} as predicted. All curves are are plotted as the mean with 95\% confidence regions shaded}
        \label{fig:multiple_single_task}
\end{figure}

\ifshort{\vspace{-0.2cm}}{}
\section{Discussion and Future Work}
\ifshort{\vspace{-0.2cm}}{}


We propose an efficient algorithm to provably recover linear operators across multiple tasks to optimality from non-iid non-isotropic data, recovering near oracle empirical risk minimization rates. We show that the benefit of learning over multiple tasks manifests in a lower noise level in the optimization and smaller sample requirements for individual tasks. These results contribute toward a general understanding of representation learning from an algorithmic and statistical perspective. Some immediate open questions are: whether good initialization of the representation is necessary, and whether the convergence rate of \texttt{DFW} can be optimized e.g., through $\ell^2$-regularized weights $\hatFh$.
Resolving these questions has important implications for the natural extension of our framework: as emphasized in \citep{collins2021exploiting}, the alternating empirical risk minimization (holding representation fixed) and gradient descent (holding task-specific weights fixed) framework naturally extends to the nonlinear setting. Providing guarantees for nonlinear function classes is an exciting and impactful avenue for future work, which concurrent work is moving toward, e.g.\ for $2$-layer ReLU networks \citep{collins2023provable} and kernel ridge regression \citep{meunier2023nonlinear}. It remains to be seen whether a computationally-efficient algorithm can be established for nonlinear meta-learning in the non-iid and non-isotropic data regime, while preserving joint scaling in number of tasks and data.







\section*{Acknowledgements}

The authors thank Stephen Tu and Ingvar Ziemann for various helpful comments. TZ and NM gratefully acknowledge support from NSF Award
SLES 2331880, NSF CAREER award ECCS 2045834, and NSF EECS 2231349. LT is funded by the Columbia Presidential Fellowship. JA is partially funded by NSF grants ECCS 2144634 and 2231350 and the Columbia Data Science Institute.

\bibliographystyle{unsrtnat}
\bibliography{refs, multitaskIL_refs}


\clearpage
\appendix

\tableofcontents
\clearpage

\section{Theoretical Analysis of \texttt{DFW} (\Cref{alg: multi-task alt min descent})} \label{sec: full theoretical analysis}

\subsection{Preliminaries}

We introduce some preliminary concepts and results that recur throughout our analysis. A fundamental concept in the analysis of least-squares solutions is the self-normalized martingale \citep{abbasi2011online, abbasi2013online}.

\begin{lemma}[cf.\ {\citet[Lemma B.3]{zhang2022multitask}}]
\label{lemma:yasin_supermartingale}
Let $\scurly{x_i}_{i \geq 1}$ be a $\R^{d_x}$-valued process adapted to a filtration $\scurly{\calF_i}_{i \geq 1}$.
Let $\scurly{w_i}_{i \geq 1}$ be a $\R^{d_y}$-valued process adapted to $\scurly{\calF_i}_{i \geq 2}$.
Suppose that $\scurly{w_i}_{i \geq 1}$ is a $\sigma^2$-subgaussian
martingale difference sequence, i.e.,:
\begin{align}
    \Ex[ w_i \mid \calF_i ] &= 0, \\
    \Ex[ \exp(\lambda  v^\top w_i ) \mid \calF_i ] &\leq \exp\paren{\frac{\lambda^2 \sigma^2 \norm{v}^2}{2}} \:\:\forall \calF_i \text{-measurable } \lambda \in \R, v \in \R^{d_y}.
\end{align}
For $\Lambda \in \R^{d_y \times d_x}$,
let $\scurly{M_k(\Lambda)}_{k \geq 1}$ be the $\R$-valued process:
\begin{align}
    M_k(\Lambda) = \exp\left( \frac{1}{\sigma}\sum_{i=1}^{k} \langle \Lambda x_i, w_i \rangle - \frac{1}{2}\sum_{i=1}^{k} \norm{\Lambda x_i}^2   \right).
\end{align}
Then, the process $\scurly{M_k(\Lambda)}_{k \geq 1}$
satisfies $\Ex[ M_k(\Lambda) ] \leq 1$ for all $k \geq 1$.
\end{lemma}

In particular, this implies the following self-normalized martingale inequality that handles multiple matrix-valued self-normalized martingales simultaneously. This can be seen as an instantiation of the Hilbert space variant from \cite{abbasi2013online}.

\begin{proposition}[cf.\ {\citet[Prop.\ B.1]{zhang2022multitask}}, {\citet[Proposition 8.2]{sarkar2021finite}}]
\label{prop:yasin_multiprocesses}
Fix $T \in \N_+$. For $t \in [T]$,
let $\scurly{x_i^{(t)}, w_i^{(t)}}_{i \geq 1}$ be a $\R^{d_x} \times \R^{d_y}$-valued process
and $\scurly{\calF_i^{(t)}}_{i \geq 1}$ be a filtration such that
$\scurly{x_i^{(t)}}_{i \geq 1}$ is adapted to $\scurly{\calF_i^{(t)}}_{i \geq 1}$,
$\scurly{w_i^{(t)}}_{i \geq 1}$ is adapted to $\scurly{\calF_i^{(t)}}_{i \geq 2}$, and 
$\scurly{w_i^{(t)}}_{i \geq 1}$ is a $\sigma^2$-subgaussian martingale difference sequence.
Suppose that for all $t_1 \neq t_2$, the process
$\scurly{x_i^{(t_1)}, w_i^{(t_1)}}$ is independent
of $\scurly{x_i^{(t_2)}, w_i^{(t_2)}}$.
Fix (non-random) positive definite matrices $\{V^{(t)}\}_{t=1}^{T}$.
For $k \geq 1$ and $t \in [T]$, define $\hat\Sigma^{(t)}_k := \sum_{i=1}^{k} x_i x_i^\top$.
Then, given any fixed $N,T \in \N_+$,
with probability at least $1-\delta$:
\begin{align}
    \sum_{t=1}^{T} \norm{\sum_{i=1}^N \whit \xhit^\top \paren{V^{(t)} + \hat\Sigma^{(t)}_N}^{-1/2} }_F^2 \leq d_y \sigma^2 \sum_{t=1}^T \log\paren{\frac{\det\paren{V^{(t)} + \hat\Sigma^{(t)}_N}}{\det(V^{(t)})}} + 2\sigma^2 \log(1/\delta)
\end{align}
Alternatively, in the spectral norm, we have with probability at least $1 - \delta$,
\begin{align}\label{eq: spectral norm SNM bound}
    \begin{split}
        \sum_{t=1}^{T} \norm{\sum_{i=1}^N \whit \xhit^\top \paren{V^{(t)} + \hat\Sigma^{(t)}_N}^{-1/2} }_2^2 &\leq 4 \sigma^2 \sum_{t=1}^T \log\paren{\frac{\det\paren{V^{(t)} + \hat\Sigma^{(t)}_N}}{\det(V^{(t)})}} \\
    &\quad + 13 d_y T \sigma^2 + 8\sigma^2 \log(1/\delta).
    \end{split}
\end{align}

\end{proposition}


We note that the above bound also holds for individual tasks $t \in [T]$ simply by removing the summand over $t$. We introduce the following useful two-sided concentration inequality for the sample covariance of iid subgaussian covariates.
\begin{lemma}[{\citet[Claim A.1, A.2]{du2020few}}]\label{lem: covariance concentration}
    Let $x_1,\dots,x_N \in \R^d$ be iid random vectors that satisfy $\Ex[x_i] = 0$, $\Ex\brac{x_i x_i^\top} = \Sigma$, and $x_i$ is $\gamma^2$-subgaussian. Fix $\delta \in (0,1)$. Suppose $N \gtrsim \gamma^4\paren{d + \log(1/\delta)}$. Then with probability at least $1 - \delta$, the following holds
    \begin{align}
        0.9 \Sigma \preceq \frac{1}{N}\sum_{i=1}^N x_i x_i^\top \preceq 1.1 \Sigma. 
    \end{align}
    Furthermore, for any matrix $U \in \R^{r \times d_x}$, as long as $N \gtrsim \gamma^4 \paren{r + \log(1/\delta)}$, we have with probability at least $1 - \delta$
    \begin{align}
        0.9 U \Sigma U^\top \preceq \frac{1}{N}\sum_{i=1}^N U x_i x_i^\top U^\top \preceq 1.1 U \Sigma U^\top.
    \end{align}
\end{lemma}

Combining \Cref{prop:yasin_multiprocesses} and \Cref{lem: covariance concentration} yields the following self-normalized martingale bound without the non-random lower bound $V^{(t)}$.
\begin{lemma}\label{lem: SNM bound with covariance conc}
    Consider the quantities defined in \Cref{prop:yasin_multiprocesses} and assume $\xhit$ are zero-mean and $\gamma^2$-subgaussian. Then, as long as $N \gtrsim \gamma^4 \paren{d_x + \log(T/\delta)}$, with probability at least $1 - \delta$:
    \begin{align*}
        \sum_{t=1}^{T} \norm{\sum_{i=1}^N \whit \xhit^\top \paren{\hat\Sigma^{(t)}_N}^{-1/2} }_F^2 &\leq 2 d_y d_x T \sigma^2  + 4\sigma^2 \log(T/\delta), \text{ or} \\
        \sum_{t=1}^{T} \norm{\sum_{i=1}^N \whit \xhit^\top \paren{\hat\Sigma^{(t)}_N}^{-1/2} }_2^2 &\leq 8 d_x T \sigma^2  + 26d_y T \sigma^2 + 16\sigma^2 \log(T/\delta).
    \end{align*}
\end{lemma}

\textit{Proof: } we observe that if $\hat\Sigma_N^{(t)} \succeq V^{(t)}$, then
\[
2 \hat\Sigma_N^{(t)} \succeq V^{(t)} + \hat\Sigma_N^{(t)} \implies \paren{\hat\Sigma_N^{(t)}}^{-1} \preceq 2\paren{V^{(t)} + \hat\Sigma_N^{(t)}}^{-1}.
\]
This implies
\begin{align*}
    &\sum_{t=1}^{T} \ones\curly{\hat\Sigma_N^{(t)} \succeq V^{(t)}} \norm{\sum_{i=1}^N \whit \xhit^\top \paren{\hat\Sigma^{(t)}_N}^{-1/2} }^2 \\
    \leq \;& 2 \sum_{t=1}^{T} \ones\curly{\hat\Sigma_N^{(t)} \succeq V^{(t)}}  \norm{\sum_{i=1}^N \whit \xhit^\top \paren{V^{(t)} + \hat\Sigma^{(t)}_N}^{-1/2} }^2.
\end{align*}
Defining $\Sigma^{(t)} := \Ex\brac{\xhit \xhit^\top}$, let us consider for each $t$ the event:
\[
0.9 \Sigma^{(t)} \preceq \hat\Sigma_N^{(t)} \preceq 1.1 \Sigma^{(t)},
\]
which by \Cref{lem: covariance concentration} occurs with probability at least $1 - \delta$ as long as $N \gtrsim \gamma^4(d_x + \log(1/\delta))$. Setting $V^{(t)} := 0.9 N \Sigma^{(t)}$ and conditioning on the above event, we observe that by definition $\hat\Sigma_N^{(t)} \succeq V^{(t)}$, and
\begin{align*}
    \log \left(\frac{\det\paren{V^{(t)} + \hat\Sigma^{(t)}_N}}{\det(V^{(t)})}\right) &= \log \det\paren{I_{d_x} + \hat\Sigma_N^{(t)}\paren{V^{(t)}}^{-1}} \\
    &= \log \det\paren{\paren{1 + \frac{1.1}{0.9}}I_{d_x} }  \\
    &\leq d_x.
\end{align*}
Plugging this into \Cref{prop:yasin_multiprocesses} and union-bounding over the desirable event over each $t \in [T]$, and adjusting the failure probability $\delta/T \mapsto \delta$, we get our desired result. $\qedhere$ 


In order to instantiate our bounds for non-iid covariates, we introduce the notions of $\beta$-mixing stationary processes \citep{kuznetsov2017generalization, tu2018least}.
\begin{definition}[$\beta$-mixing]
Let $\scurly{x_i}_{t \geq 1}$ be a $\R^d$-valued discrete-time stochastic process adapted to filtration $\scurly{\calF_i}_{t = 1}^\infty$. 
We denote the stationary distribution $\nu_\infty$. We define the $\beta$-mixing coefficient
\begin{equation}
    \beta(k) := \sup_{t \geq 1} \Ex_{\curly{x_\ell}_{\ell=1}^t}\brac{\norm{\prob_{x_{t+k}}(\cdot \mid \calF_i) - \nu_\infty }_{\mathrm{tv}}},
\end{equation}
where $\norm{\cdot}_{\mathrm{tv}}$ denotes the total variation distance between probability measures.
\end{definition}
Intuitively, the $\beta$-mixing coefficient measures how quickly on average a process mixes to the stationary distribution along any sample path. To see how $\beta$-mixing is instantiated, let $\curly{x_i}_{t=1}^T$ be a sample path from a $\beta$-mixing process. Consider the following subsampled paths formed by taking every $a$-th covariate of $\curly{x_i}$:
\begin{equation}
    X^T_{(j)} := \curly{x_i: 1 \leq t \leq T, \; (t-1 \;\mathrm{mod}\;a) = j-1}, \;\; j=1,\dots,a.
\end{equation}
Let the integers $m_1,\dots, m_a$ and index sets $I_{(1)},\dots I_{(a)}$ denote the sizes and indices of $X^T_{(1)},\dots, X^T_{(a)}$, respectively. Finally, let $X^{m_j}_\infty$ denote a sequence of $m_j$ iid draws from the stationary distribution $\nu_\infty$. The following is a key lemma in relating a correlated process to iid draws.
\begin{lemma}[{\citet[Proposition 2]{kuznetsov2017generalization}}]\label{lem: kuznetsov mixing}
    Let $g(\cdot)$ be a real-valued Borel-measurable function satisfying $-M_1 \leq g(\cdot) \leq M_2$ for some $M_1,M_2 \geq 0$. Then, for all $j = 1,\dots, a$.
    \begin{align*}
        \abs{\Ex\sbrac{g(X_\infty^{m_j})} - \Ex\sbrac{g(X_{(j)}^T)}  } \leq (M_1 + M_2) m_j \beta(a).
    \end{align*}
\end{lemma}
In our analysis, we often instantiate \Cref{lem: kuznetsov mixing} with $g(\cdot)$ as an indicator function on a success event. For appropriately selected block length $a$, we are thus able to relate simpler iid analysis on $X_\infty^{m_j}$ to the original process $X^T_{(j)}$, accruing an additional factor in the failure probability. Lastly, we introduce a standard matrix concentration inequality.
\begin{lemma}[Matrix Hoeffding \citep{tropp2011user}]\label{lem: matrix hoeffding}
    Let $\scurly{X_t}_{t=1}^T \subset \R^{d \times d}$ be a sequence of independent, random symmetric matrices, and let $\scurly{B_t}_{t=1}^T$ be a sequence of fixed symmetric matrices. Assume each random matrix satisfies
    \[
    \Ex\brac{X_t}  = 0, \quad X_t^2 \preceq B_t^2 \text{ almost surely}.
    \]
    Then for all $t \geq 0$,
    \[
    \prob\brac{\lambda_{\max}\paren{\sum_{t=1}^T X_t} \geq t} \leq d\cdot \exp\paren{-\frac{t^2}{8\sigma^2}}, \quad \sigma^2 := \norm{\sum_{t=1}^T B_t^2}.
    \]
    In particular, for general rectangular $\scurly{M_t}_{t=1}^T \subset \R^{d_1 \times d_2}$, we may define $X_t := \bmat{0 & M_t \\ M_t^\top & 0}$ to yield a singular value concentration inequality. Assume each $M_t$ satisfies
    \[
    \Ex\brac{M_t}  = 0, \quad X_t^2 \preceq B_t^2 \text{ almost surely}.
    \]
    Then for all $t \geq 0$,
    \[
    \prob\brac{\sigma_{\max}\paren{\sum_{t=1}^T M_t} \geq t} \leq (d_1 + d_2)\cdot \exp\paren{-\frac{t^2}{8\sigma^2}}, \quad \sigma^2 := \norm{\sum_{t=1}^T B_t^2}.
    \]
\end{lemma}
As hinted by the indexing of the matrices, by leveraging the independence of processes across tasks $t$, \Cref{lem: matrix hoeffding} can be used to bound various quantities averaged across tasks, under the important caveat that the matrices are \textit{zero-mean}, which ties back to the necessity of our de-biasing and feature-whitening adjustments.



\subsection{The IID Setting}

We recall that given the current representation iterate $\hat\Phi$, an iid draw of a multitask dataset $\scurly{(\xhit, \yhit)}_{t=1, i=1}^{T, N}$, $t = 1,\dots, T$, and \texttt{DFW} trajectory partitions $\calN_1,\calN_2$, the least squares weights $\hatFh$ can be written as
\begin{align}
    \hat F^{(t)} &= \argmin_{F} \sum_{i\in\calN_1} \norm{\yhit - F\zhit}^2 \nonumber\\
    &= \Fh_\star \Phi_\star \Xhone^\top \Zhone \paren{\hat\Sigma^{(t)}_{z, \calN_1}}^{-1} + \Whone^\top \Zhone \paren{\hat\Sigma^{(t)}_{z, \calN_1}}^{-1} \nonumber \\
    &= \Fhstar \Phi_\star \hat\Phi^\top + \Fhstar \Phi_\star  \paren{I_{d_x} - \hat\Phi^\top \hat\Phi} \Xhone^\top \Zhone\paren{\hat\Sigma^{(t)}_{z, \calN_1}}^{-1} + \Whone^\top \Zhone \paren{\hat\Sigma^{(t)}_{z, \calN_1}}^{-1}. \label{eq: weight update}
\end{align}
Now recalling the \texttt{DFW} representation update in the iid setting \eqref{eq: update rule FW IID case}, we have
\begin{align}\label{eq: representation update}
    R \hat\Phi_+ &= \hat\Phi - \frac{\eta}{T}\sum_{t=1}^T \mathopen{}\hat F^{(t)}\mathclose{}^\top \paren{\hat F^{(t)} \hat\Phi - \Fh_\star \Phi_\star} - \frac{\eta}{T}\sum_{t=1}^T \mathopen{}\hat F^{(t)}\mathclose{}^\top \Whtwo^\top \Xhtwo \paren{\hatSigmaxhNT}^{-1}.
\end{align}
Right multiplying the update by $\Phiperp^\top$, we get
\begin{align*}
    R \hat\Phi_+ \Phiperp^\top &= \hat\Phi \Phiperp^\top  - \frac{\eta}{T}\sum_{t=1}^T \mathopen{}\hat F^{(t)}\mathclose{}^\top \paren{\hat F^{(t)} \hat\Phi - \Fh_\star \Phi_\star}\Phiperp^\top - \frac{\eta}{T}\sum_{t=1}^T \mathopen{}\hat F^{(t)}\mathclose{}^\top \Whtwo^\top \Xhtwo \paren{\hat\Sigma^{(t)}_{x, \calN_2}}^{-1} \Phiperp^\top \\
    &= \underbrace{\paren{I_{dx} - \frac{\eta}{T}\sum_{t=1}^T \mathopen{}\hat F^{(t)}\mathclose{}^\top\hat F^{(t)}}  \hat\Phi \Phiperp^\top}_{\text{``contraction'' term}} - \underbrace{\frac{\eta}{T}\sum_{t=1}^T \mathopen{}\hat F^{(t)}\mathclose{}^\top \Whtwo^\top \Xhtwo \paren{\hat\Sigma^{(t)}_{x, \calN_2}}^{-1} \Phiperp^\top}_{\text{``noise'' term}},
\end{align*}
where the last line is composed of a contraction term and a noise term. We start with an analysis of the noise term. 

\subsection*{Bounding the noise term}\label{appdx: noise term bound}
We observe that since $\hatFh$ is by construction independent of $\Whtwo, \Xhtwo$, by the independence of $\xhit$ across $t$ and $i$, and the noise independence $\whit \perp \xhit$, we find
\begin{align*}
    \Ex\brac{\frac{1}{T}\sum_{t=1}^T \mathopen{}\hat F^{(t)}\mathclose{}^\top \Whtwo^\top \Xhtwo \paren{\hat\Sigma^{(t)}_{x, \calN_2}}^{-1}} &= \frac{1}{T}\sum_{t=1}^T \Ex\brac{\hatFh}^\top \Ex\brac{\Whtwo}\Ex\brac{\Xhtwo \paren{\hat\Sigma^{(t)}_{x, \calN_2}}^{-1}} = 0.
\end{align*}
Therefore, we set up for an application of \Cref{lem: matrix hoeffding}. Toward doing so, we prove the following two ingredients: 1.\ a high probability bound on $\snorm{\hatFh}$, 2.\ a high probability bound on the least-squares noise-esque term $\snorm{\mathopen{}\hat F^{(t)}\mathclose{}^\top \Whtwo^\top \Xhtwo \paren{\hat\Sigma^{(t)}_{x, \calN_2}}^{-1}}$. We then condition on these two high-probability events to instantiate the almost-sure boundedness in \Cref{lem: matrix hoeffding}. We start with the analysis of $\hatFh$. By \eqref{eq: weight update}, we have
\begin{align*}
    \norm{\hatFh} &\leq \norm{\Fhstar} + \norm{\Fhstar} \norm{\Phi_\star \calP_{\hat\Phi}^\perp \Xhone^\top \Zhone\paren{\hat\Sigma^{(t)}_{z, \calN_1}}^{-1}} + \norm{\Whone^\top \Zhone \paren{\hat\Sigma^{(t)}_{z, \calN_1}}^{-1}}.
\end{align*}
\begin{lemma}\label{lem: weight term concentration}
    Let $\abs{\calN_1} := N_1 \gtrsim \gamma^4\paren{r + \log(1/\delta)}$. Then, with probability greater than $1 - \delta$, we have
    \begin{align}
        \norm{\Phi_\star \calP_{\hat\Phi}^\perp  \Xhone^\top \Zhone\paren{\hat\Sigma^{(t)}_{z, \calN_1}}^{-1}} &\leq \frac{5}{4}\dist(\hat\Phi, \Phi_\star) \kappa\mem\paren{\Sigmaxh}, \label{eq: misspecification term bound} \\
        \norm{\Whone^\top \Zhone \paren{\hat\Sigma^{(t)}_{z, \calN_1}}^{-1}} &\lesssim \sigmahw \sqrt{\frac{d_y + r + \log(1/\delta)}{\lambda_{\min}(\Sigmazh) N_1 }}. \label{eq: weights SNM bound} 
    \end{align}
\end{lemma}
\textit{Proof: } we begin with the bound \eqref{eq: misspecification term bound}. We observe that by definition $\frac{1}{N_1} \Xhone^\top \Xhone = \hat\Sigma^{(t)}_{x, \calN_1}$, and $\Phi_\star \calP_{\hat\Phi}^\perp, \hat\Phi$ are $r \times d_x$ matrices. Therefore, we invoke \Cref{lem: covariance concentration} twice to find: as long as $N_1 \gtrsim \gamma^4 \paren{r + \log(1/\delta)}$, with probability at least $1 - \delta$, the following bounds hold simultaneously:
\begin{align*}
    \frac{1}{N_1} \norm{\Phi_\star \calP_{\hat\Phi}^\perp  \Xhone^\top}^2 &\leq 1.1 \norm{\Phi_\star \calP_{\hat\Phi}^\perp \paren{\Sigmaxh}^{1/2}}^2 \leq 1.1 \dist(\hat\Phi, \Phi_\star)^2 \lambda_{\max}\paren{\Sigmaxh} \\
    \frac{1}{N_1} \norm{\Zhone}^2 &\leq 1.1 \norm{\Sigmazh} \leq 1.1 \lambda_{\max}\paren{\Sigmaxh} \\
    \norm{\paren{\hat\Sigma^{(t)}_{z, \calN_1}}^{-1}} &\leq 0.9 \lambda_{\min}(\Sigmazh)^{-1} \leq 0.9 \lambda_{\min}(\Sigmaxh)^{-1},
\end{align*}
where we recall that $\Sigmazh = \hat\Phi \Sigmaxh \hat\Phi^\top$. Therefore, applying Cauchy-Schwarz on the LHS of \eqref{eq: misspecification term bound} and the above bounds (converting $1.1/0.9 < 5/4$) yields the desired upper bound on the RHS. Moving onto \eqref{eq: weights SNM bound}, we observe that since $\hat\Phi$ is fixed, $\scurly{\whit, \zhit}_{i\geq 1}$ is an $\R^{d_y} \times \R^r$-valued martingale difference sequence. Therefore, we may apply \Cref{lem: covariance concentration} and \Cref{lem: SNM bound with covariance conc} to find: as long as $N_1 \gtrsim \gamma^4 \paren{r + \log(1/\delta)}$, with probability at least $1 - \delta$,
\begin{align*}
    \norm{\Whone^\top \Zhone \paren{\hat\Sigma^{(t)}_{z, \calN_1}}^{-1}} &\leq  \norm{\Whone^\top \Zhone \paren{\hat\Sigma^{(t)}_{z, \calN_1}}^{-1/2}} \lambda_{\min}\paren{\hat\Sigma^{(t)}_{z, \calN_1}}^{-1/2} \\
    &\lesssim \sqrt{\frac{\sigmahw^2 \paren{d_y + r + \log(1/\delta)} }{\lambda_{\min}(\Sigmaxh)N_1}},
\end{align*}
which establishes the bound \eqref{eq: weights SNM bound}.
$\qedhere$

Therefore, by assuming $\dist(\hat\Phi, \Phi_\star)$ is sufficiently small, and that $N_1$ is large enough to offset the noise bound \eqref{eq: weights SNM bound}, we immediately get the following bound relating $\snorm{\hatFh}$ to $\snorm{\Fhstar}$.
\begin{lemma}\label{lem: hp bound on hatFh}
    Assume
    \begin{align*}
        N_1 \gtrsim \max\curly{\gamma^4 \paren{r + \log(1/\delta)}, \frac{\sigmahw^2}{C^2\snorm{\Fhstar}^2\lambda_{\min}(\Sigmaxh)} \paren{d_y + r + \log(1/\delta)}},
    \end{align*}
    and $\dist(\hat\Phi, \Phi_\star) \leq \frac{2}{5}C \kappa\mem\paren{\Sigmaxh}^{-1}$, for fixed $C > 0$. Then with probability at least $1 - \delta$
    \begin{align}
        \snorm{\hatFh} &\leq (1+C) \snorm{\Fhstar}.
    \end{align}
\end{lemma}
The factor $C > 0$ serves as a free parameter which we can determine later--a larger $C$ implies a relaxed requirement on the initial condition (disappears when $C \geq \frac{5}{2} \kappa(\Sigmaxh)$) and burn-in requirement on $N_1$, but results in a larger subgaussian-parameter bound on the representation update noise term \eqref{eq: representation update}, as we demonstrate: given that the success event of \Cref{lem: hp bound on hatFh} holds, then by the definition of subgaussianity (\Cref{assumption: subgaussian covariate}), we observe that $\hatFh^\top \whit$ is zero-mean and $(1+C)^2 \sigmahw^2\|\Fhstar\|^2$-subgaussian, supported on $\R^r$. Therefore, bounding
\[
\norm{\mathopen{}\hat F^{(t)}\mathclose{}^\top \Whtwo^\top \Xhtwo \paren{\hat\Sigma^{(t)}_{x, \calN_2}}^{-1}} \leq \norm{\hatFh\Whtwo^\top\Xhtwo \paren{\hat\Sigma^{(t)}_{x, \calN_2}}^{-1/2}} \lambda_{\min}(\hat\Sigma^{(t)}_{x, \calN_2})^{-1/2},
\]
we invoke \Cref{prop:yasin_multiprocesses} and \Cref{lem: covariance concentration} to get the following bound.
\begin{lemma}\label{lem: task-wise SNM bound}
    Let the conditions of \Cref{lem: hp bound on hatFh} hold. Then, for a fixed $t \in [T]$, as long as $\abs{\calN_2} := N_2 \gtrsim \gamma^4 \paren{d_x + \log(1/\delta)}$, with probability at least $1 - \delta$,
    \begin{align}
        \norm{\mathopen{}\hat F^{(t)}\mathclose{}^\top \Whtwo^\top \Xhtwo \paren{\hat\Sigma^{(t)}_{x, \calN_2}}^{-1}} &\lesssim (1+C) \sigmahw \|\Fh_\star\| \sqrt{\frac{d_x + \log(1/\delta)}{\lambda_{\min}(\Sigmaxh) N_2}}.
    \end{align}
\end{lemma}
With a bound on the task-specific noise term in hand, we may now produce the final bound on the (task-averaged) noise term.
\begin{proposition}[Noise term bound]\label{prop: DFW update noise term final bound}
    Assume
    \begin{align*}
        N_1 &\gtrsim \max\curly{\gamma^4 \paren{r + \log(T/\delta)}, \max_t \frac{\sigmahw^2}{C^2\snorm{\Fhstar}^2\lambda_{\min}(\Sigmaxh)} \paren{d_y + r + \log(T/\delta)}}, \\
        N_2 &\gtrsim \gamma^4 \paren{d_x + \log(T/\delta)},
    \end{align*}
    and $\dist(\hat\Phi, \Phi_\star) \leq \max_t \frac{2}{5}C \kappa\mem\paren{\Sigmaxh}^{-1}$, for fixed $C > 0$. Then, with probability at least $1 - \delta$,
    \begin{align*}
        \norm{\frac{1}{T}\sum_{t=1}^T \mathopen{}\hat F^{(t)}\mathclose{}^\top \Whtwo^\top \Xhtwo \paren{\hat\Sigma^{(t)}_{x, \calN_2}}^{-1}} &\lesssim
        \sigmaavg (1+C) \sqrt{\frac{d_x + \log(T/\delta)}{TN_2} \log\paren{\frac{d_x}{\delta}}},
    \end{align*}
    where $\sigmaavg := \sqrt{\frac{1}{T}\sumT \frac{\sigmahw^2 \snorm{\Fhstar}^2}{\lambda_{\min}(\Sigmaxh)}}$ is the \emph{task-averaged} noise-level.
\end{proposition}

\textit{Proof of \Cref{prop: DFW update noise term final bound}:} we set up for an application of the Matrix Hoeffding bound (\Cref{lem: matrix hoeffding}). By union bounding over the task-specific noise bound \Cref{lem: task-wise SNM bound}, we have with probability at least $1 - \delta$, for all $t \in [T]$ simultaneously:
\begin{align*}
    \norm{\mathopen{}\hat F^{(t)}\mathclose{}^\top \Whtwo^\top \Xhtwo \paren{\hat\Sigma^{(t)}_{x, \calN_2}}^{-1}} &\lesssim (1+C) \frac{\sigmahw \|\Fh_\star\|}{\sqrt{\lambda_{\min}(\Sigmaxh)}} \sqrt{\frac{d_x + \log(T/\delta)}{N_2}},
\end{align*}
given the assumed burn-in conditions hold. Therefore, by setting
\begin{align*}
    B^{(t)} &=  \calO(1) \frac{\sigmahw \|\Fh_\star\|}{\sqrt{\lambda_{\min}(\Sigmaxh)}} \sqrt{\frac{d_x + \log(T/\delta)}{N_2}} I_{d_x + r} \\
    \sigma^2 &= \norm{\sumT (B^{(t)})^2} = \paren{ \sumT \frac{\sigmahw^2 \|\Fh_\star\|^2}{\lambda_{\min}(\Sigmaxh)}}(1+C)^2 \frac{d_x + \log(T/\delta)}{N_2} \\
    &= T \sigmaavg^2(1+C)^2 \frac{d_x + \log(T/\delta)}{N_2}
\end{align*}
we invoke \Cref{lem: matrix hoeffding} and invert
\[
\paren{d_x + r} \cdot \exp\paren{\frac{-t^2}{8\sigma^2}} \leq \delta
\]
to set
\begin{align*}
    t \approx \sqrt{T}\sigmaavg (1+C) \sqrt{\frac{d_x + \log(T/\delta)}{N_2} \log\paren{\frac{d_x}{\delta}}}, \quad \text{recalling }r \leq d_x.
\end{align*}
The resulting Hoeffding bound yields
\begin{align*}
    & \prob\brac{ \norm{\sumT \mathopen{}\hat F^{(t)}\mathclose{}^\top \Whtwo^\top \Xhtwo \paren{\hat\Sigma^{(t)}_{x, \calN_2}}^{-1} }\gtrsim  \sqrt{T}\sigmaavg (1+C) \sqrt{\frac{d_x + \log(T/\delta)}{N_2} \log\paren{\frac{d_x}{\delta}}} } \leq \delta  \\
    \iff \;& \prob\brac{ \norm{\frac{1}{T}\sumT \mathopen{}\hat F^{(t)}\mathclose{}^\top \Whtwo^\top \Xhtwo \paren{\hat\Sigma^{(t)}_{x, \calN_2}}^{-1} }\gtrsim  \sigmaavg (1+C) \sqrt{\frac{d_x + \log(T/\delta)}{TN_2} \log\paren{\frac{d_x}{\delta}}} } \leq \delta
\end{align*}
$\qedhere$

We have bounded the noise term on the DFW representation update, demonstrating critically the bound on the noise term benefits from a scaling with the number of tasks $T$. The task-relevant quantity $\sigmaavg$ quantifies that the ``noise level'' of the problem is an \textit{average} over the noise-levels of each task. We note that our application of Matrix Hoeffding is rather crude, and the above bound can likely be improved in terms of polylog$(1/\delta)$ factors with stronger moment bounds on the matrix-valued self-normalized martingale terms, but this is out of the scope of this paper. 

Returning to the choice of $C$, $C \approx 1$ implies no further system/task-specific dependence beyond the terms in $K^{(1:T)}$; however, this may translate into a stringent requirement on the burn-in $N_1$ and the subspace distance $\dist(\hat\Phi, \Phi_\star)$. On the other hand, $C \gtrsim \sqrt{\kappa\mem(\Sigmaxh)}$ relaxes the burn-in and potentially renders the subspace distance requirement trivial, but manifests a condition number in the noise bound. We note that as we expect $\dist(\hat\Phi, \Phi_\star)$ to decrease geometrically with iterations of DFW, the subspace distance requirement is only relevant for the first few iterations. In general, this intuitively captures the cost of ill-conditioned data distributions. We now move on to bounding the contraction term.

\subsection*{Bounding the Contraction Term}

Let us define,
\begin{align*}
    \Deltah &:= \Fhstar \Phi_\star  \paren{I_{d_x} - \hat\Phi^\top \hat\Phi} \Xhone^\top \Zhone\paren{\hat\Sigma^{(t)}_{z, \calN_1}}^{-1}\\
    \Eh &:= \Whone^\top \Zhone \paren{\hat\Sigma^{(t)}_{z, \calN_1}}^{-1},
\end{align*}
such that we may write \eqref{eq: weight update} as $\hatFh = \Fhstar \Phi_\star \hat\Phi^\top + \Deltah + \Eh$. We expand
\begin{align*}
    \hatFh^\top \hatFh &= \hat\Phi\Phi_\star^\top \Fhstar^\top \Fhstar \Phi_\star \hat\Phi^\top + \Deltah^\top \Deltah + \Eh^\top \Eh  \\
    &\quad + \Sym(\Deltah^\top \Fhstar \Phi_\star \hat\Phi^\top) + \Sym(\Eh^\top \Fhstar \Phi_\star \hat\Phi^\top) + \Sym(\Deltah^\top \Eh),
\end{align*}
where $\Sym(A) := A + A^\top$.
We will make repeated use of the following matrix Cauchy-Schwarz-type lemma.
\begin{lemma}\label{lem: matrix cauchy-schwarz}
Let $A_t, B_t$ be real-valued matrices for $t = 1,\dots,T$. Then,
    \begin{align*}
        \norm{\sum_{t=1}^T A_t B_t } &\leq \norm{\sum_{t=1}^T A_t A_t^\top}^{1/2} \norm{\sum_{t=1}^T B_t^\top B_t}^{1/2}.
    \end{align*}
\end{lemma}
Now taking the average over tasks $T$, we then may write
\begin{align*}
    &\lambda_{\max}\paren{\frac{1}{T}\sumT \hatFh^\top \hatFh} \\
    \leq&\; \lambda_{\max}\paren{\frac{1}{T}\sumT \Fhstar^\top \Fhstar} + \lambda_{\max}\paren{\frac{1}{T}\sumT\Deltah^\top \Deltah} + \lambda_{\max}\paren{\frac{1}{T}\sumT \Eh^\top \Eh} \\
    &+ \norm{\Sym(\frac{1}{T}\sumT \Deltah^\top \Fhstar \Phi_\star \hat\Phi^\top)} + \norm{\Sym(\frac{1}{T}\sumT \Eh^\top \Fhstar \Phi_\star \hat\Phi^\top)} + \norm{\Sym(\frac{1}{T}\sumT \Deltah^\top \Eh)}.
\end{align*} 
We observe that
\begin{align*}
    \norm{\Sym(A)} &= \max_{\norm{u},\norm{v}=1} u^\top A v + u^\top A^\top v \\
    &\leq 2\norm{A},
\end{align*}
and thus applying the above fact and \Cref{lem: matrix cauchy-schwarz} on the cross terms yields
\begin{align*}
    &\lambda_{\max}\paren{\frac{1}{T}\sumT \hatFh^\top \hatFh} \\
    \leq&\; \lambda_{\max}\paren{\frac{1}{T}\sumT \Fhstar^\top \Fhstar} + \lambda_{\max}\paren{\frac{1}{T}\sumT\Deltah^\top \Deltah} + \lambda_{\max}\paren{\frac{1}{T}\sumT \Eh^\top \Eh} \\
    & + 2\norm{\frac{1}{T}\sumT \Deltah^\top \Deltah}^{1/2} \norm{\frac{1}{T}\sumT \Fhstar^\top \Fhstar}^{1/2} + 2\norm{\frac{1}{T}\sumT \Eh^\top \Eh}^{1/2} \norm{\frac{1}{T}\sumT \Fhstar^\top \Fhstar}^{1/2} \\
    & + 2\norm{\frac{1}{T}\sumT \Deltah^\top \Deltah}^{1/2} \norm{\frac{1}{T}\sumT \Eh^\top \Eh}^{1/2}.
\end{align*}
Using \Cref{lem: weight term concentration}, we get the following upper bound on $\lambda_{\max}\paren{\frac{1}{T}\sumT \hatFh^\top \hatFh}$.
\begin{lemma}\label{lem: weights upper bound}
    Let
    \begin{align*}
    N_1 &\gtrsim \max\curly{\gamma^4\paren{r + \log(T/\delta)},  \frac{\bfFmax^{-1}}{c_2 T}\sumT \frac{\sigmahw^2 (d_y + r + \log(T/\delta))}{\lambda_{\min}(\Sigmaxh) } },
    \end{align*}
    and $\dist(\hat\Phi, \Phi_\star) \leq \max_t \frac{4}{5} c_1 \kappa\mem\paren{\Sigmaxh}^{-1}$ for given constants $c_1,c_2 \in (0,1)$. Then, with probability at least $1 - \delta$, we have
    \begin{align*}
        \lambda_{\max}\paren{\frac{1}{T}\sumT \hatFh^\top \hatFh} &\leq (1+ 2c_1 + 2c_2 + (c_1 + c_2)^2)\bfFmax.
    \end{align*}
\end{lemma}
\textit{Proof:} we see that it suffices to establish $\norm{\frac{1}{T}\sumT \Deltah^\top \Deltah} \lesssim \norm{\frac{1}{T}\sumT \Fhstar^\top \Fhstar} =: \bfFmax$ and $\norm{\frac{1}{T}\sumT \Eh^\top \Eh}\lesssim \bfFmax$ in order to establish
\begin{align*}
    \lambda_{\max}\paren{\frac{1}{T}\sumT \hatFh^\top \hatFh} &\lesssim \bfFmax.
\end{align*}
We recall that 
\begin{align*}
    \Deltah &:= \Fhstar \Phi_\star  \paren{I_{d_x} - \hat\Phi^\top \hat\Phi} \Xhone^\top \Zhone\paren{\hat\Sigma^{(t)}_{z, \calN_1}}^{-1}\\
    \Eh &:= \Whone^\top \Zhone \paren{\hat\Sigma^{(t)}_{z, \calN_1}}^{-1},
\end{align*}
which by \Cref{lem: weight term concentration} admit the following high-probability bounds: let $\abs{\calN_1} := N_1 \gtrsim \gamma^4\paren{r + \log(T/\delta)}$. Then, with probability greater than $1 - \delta$, we have for each $t \in [T]$
\begin{align*}
    \norm{\Phi_\star \calP_{\hat\Phi}^\perp  \Xhone^\top \Zhone\paren{\hat\Sigma^{(t)}_{z, \calN_1}}^{-1}} &\leq \frac{5}{4}\dist(\hat\Phi, \Phi_\star) \kappa\mem\paren{\Sigmaxh} \\
    \norm{\Whone^\top \Zhone \paren{\hat\Sigma^{(t)}_{z, \calN_1}}^{-1}} &\lesssim \sigmahw \sqrt{\frac{d_y + r + \log(T/\delta)}{\lambda_{\min}(\Sigmaxh) N_1 }}.
\end{align*}
In particular, this implies
\begin{align*}
    \frac{1}{T}\sumT \Deltah^\top \Deltah \preccurlyeq \paren{\frac{1}{T}\sumT \Fhstar \Fhstar} \cdot \max_t \; \paren{\frac{5}{4}\dist(\hat\Phi, \Phi_\star) \kappa\mem\paren{\Sigmaxh}}^2 \\
    \norm{\frac{1}{T}\Eh^\top \Eh} \leq \paren{\frac{1}{T} \sumT \frac{\sigmahw^2}{\lambda_{\min}(\Sigmaxh)}}\frac{d_y + r + \log(T/\delta)}{ N_1 }
\end{align*}
It therefore suffices to set bounds on $\dist(\hat\Phi, \Phi_\star)$ and $N_1$ such that $\max_t \frac{5}{4}\dist(\hat\Phi, \Phi_\star) \kappa\mem\paren{\Sigmaxh} \leq c_1$ and $\norm{\frac{1}{T}\Eh^\top \Eh} \leq c^2_2  \norm{\frac{1}{T}\sumT \Fhstar^\top \Fhstar}$ for appropriate numerical constants $c_1, c_2$. Given
\begin{align*}
    \dist(\hat\Phi, \Phi_\star) &\leq \max_t \frac{4}{5} c_1 \kappa\mem\paren{\Sigmaxh}^{-1} \\
    N_1 &\gtrsim \max\curly{\gamma^4\paren{r + \log(T/\delta)},  \frac{\bfFmax^{-1}}{c_2^2 T}\sumT \frac{\sigmahw^2 (d_y + r + \log(T/\delta))}{\lambda_{\min}(\Sigmaxh) } },
\end{align*}
we have
\begin{align*}
    \lambda_{\max}\paren{\frac{1}{T}\sumT \hatFh^\top \hatFh}
    \leq \paren{1 + 2c_1 + 2c_2 + (c_1 + c_2)^2} \lambda_{\max}\paren{\frac{1}{T}\sumT \Fhstar^\top \Fhstar} 
\end{align*}

$\qedhere$

\Cref{lem: weights upper bound} informs the maximum we may set the step-size. To now upper bound the contraction rate, we lower bound $\lambda_{\min}(\hatFh^\top \hatFh)$. We observe that the diagonal terms $\Deltah^\top \Deltah$, $\Eh^\top \Eh$ are psd, and thus can be ignored in the lower bound. We then observe that by Weyl's inequality \citep{horn2012matrix}, we have
\begin{align*}
    \lambda_{\min}(\frac{1}{T}\sumT \hatFh^\top \hatFh) &\geq \lambda_{\min}\paren{\frac{1}{T}\sumT \hat\Phi\Phi_\star^\top \Fhstar^\top \Fhstar \Phi_\star \hat\Phi^\top} \\
    &\quad - \lambda_{\max}\paren{\frac{1}{T}\sumT \Sym(\Deltah^\top \Fhstar \Phi_\star \hat\Phi^\top) + \Sym(\Eh^\top \Fhstar \Phi_\star \hat\Phi^\top) + \Sym(\Deltah^\top \Eh)}.
\end{align*}
Just as in the upper bound of $\lambda_{\max}(\frac{1}{T}\sumT\hatFh^\top \hatFh)$, it suffices to upper bound the cross terms. Therefore, following the same proof as \Cref{lem: weights upper bound}, we have the analogous result.
\begin{lemma}\label{lem: weights lower bound}
    Let
    \begin{align*}
        N_1 &\gtrsim \max\curly{\gamma^4\paren{r + \log(T/\delta)},  \frac{\bfFmin^{-1}}{b_2^2}\frac{1}{T}\sumT \frac{\sigmahw^2 (d_y + r + \log(T/\delta))}{\lambda_{\min}(\Sigmaxh) } },
    \end{align*}
    and $\dist(\hat\Phi, \Phi_\star) \leq \frac{4}{5}b_1 \sqrt{\frac{\bfFmin}{\bfFmax}} \max_t \kappa\mem\paren{\Sigmaxh}^{-1}$ for given constants $b_1,b_2 \in (0,1)$. Then, with probability at least $1 - \delta$, we have
    \begin{align*}
        \lambda_{\min}\paren{\frac{1}{T}\sumT \hatFh^\top \hatFh} &\geq \paren{(1 - \frac{2}{3}b_1^2) - 2\paren{b_1 + b_2 + b_1 b_2}}\bfFmin.
    \end{align*}
\end{lemma}
\textit{Proof:} as in \Cref{lem: weights upper bound}, we invert the bounds from \Cref{lem: weight term concentration} for our desired factors of $\bfFmin$. In particular, observing
\begin{align*}
    \norm{\Phi_\star \calP_{\hat\Phi}^\perp  \Xhone^\top \Zhone\paren{\hat\Sigma^{(t)}_{z, \calN_1}}^{-1}} &\leq \frac{5}{4}\dist(\hat\Phi, \Phi_\star) \kappa\mem\paren{\Sigmaxh} \\
    \norm{\frac{1}{T}\sumT\Whone^\top \Zhone \paren{\hat\Sigma^{(t)}_{z, \calN_1}}^{-1}} &\lesssim \frac{1}{T}\sumT \sigmahw \sqrt{\frac{d_y + r + \log(1/\delta)}{\lambda_{\min}(\Sigmaxh) N_1 }},
\end{align*}
we invert the RHS' for $b_1 \sqrt{\frac{\bfFmin}{\bfFmax}}$ and $b_2 \sqrt{\bfFmin}$, respectively, to yield our proposed burn-in and the following guarantee with probability at least $1 - \delta$
\begin{align*}
    &\lambda_{\max}\paren{\Sym(\Deltah^\top \Fhstar \Phi_\star \hat\Phi^\top) + \Sym(\Eh^\top \Fhstar \Phi_\star \hat\Phi^\top) + \Sym(\Deltah^\top \Eh)} \\
    \leq\;& 2(b_1 + b_2 + b_1b_2)\bfFmin.
\end{align*}
The only additional factor we account for here is the lower bound on $\lambda_{\min}\paren{\frac{1}{T}\sumT \hat\Phi \Phi_\star^\top \Fhstar^\top \Fhstar \Phi_\star \hat\Phi^\top}$. We have
\begin{align*}
    \lambda_{\min}\paren{\frac{1}{T}\sumT \hat\Phi \Phi_\star^\top \Fhstar^\top \Fhstar \Phi_\star \hat\Phi^\top} &= \min_{\norm{x} = 1} x^\top \hat\Phi \Phi_\star^\top \paren{\frac{1}{T}\sumT \Fhstar^\top \Fhstar} \Phi_\star \hat\Phi^\top x \\
    &\geq \lambda_{\min}\paren{\frac{1}{T}\sumT \Fhstar^\top \Fhstar} \min_{\norm{x} = 1} x^\top \hat\Phi \Phi_\star^\top \Phi_\star \hat\Phi^\top x \\
    &= \lambda_{\min}\paren{\frac{1}{T}\sumT\Fhstar^\top \Fhstar} \sigma_{\min}^2\paren{\Phi_\star \hat\Phi^\top}.
\end{align*}
To further lower bound $\sigma_{\min}^2\paren{\Phi_\star \hat\Phi^\top}$, we observe that
\begin{align*}
    \hat\Phi \hat\Phi^\top &= \hat\Phi \paren{\Phi_\star^\top \Phi_\star + \Phiperp^\top \Phiperp} \hat\Phi^\top \\
    \implies 1 = \lambda_{\max}(\hat\Phi \hat\Phi^\top) &\leq \lambda_{\min}\paren{\hat\Phi \Phi_\star^\top \Phi_\star \hat\Phi^\top} + \lambda_{\max}\paren{\hat\Phi \Phiperp^\top \Phiperp \hat\Phi^\top} \quad \text{Weyl's inequality} \\
    \implies \sigma_{\min}^2\paren{\Phi_\star \hat\Phi^\top} &\geq 1 - \norm{\Phiperp \hat\Phi^\top}^2 =: 1 - \dist(\hat\Phi, \Phi_\star)^2.
\end{align*}
Under our assumption ensuring $\dist(\hat\Phi, \Phi_\star) \leq \frac{4}{5} b_1 \sqrt{\frac{\bfFmin}{\bfFmax}} \max_t \kappa\mem\paren{\Sigmaxh}^{-1} \leq \frac{4}{5} b_1$, we can piece together this bound with the bound on the cross terms to yield with probability at least $1 - \delta$
\begin{align*}
   \lambda_{\min}\paren{\frac{1}{T}\sumT\hatFh^\top \hatFh} &\geq \paren{(1 - \frac{2}{3}b_1^2) - 2\paren{b_1 + b_2 + b_1 b_2}}\lambda_{\min}\paren{\frac{1}{T}\sumT \Fhstar^\top \Fhstar},
\end{align*}
which completes the proof.  $\qedhere$

Piecing together \Cref{lem: weights upper bound} and \Cref{lem: weights lower bound}, and observing that if $b_1 = c_1$, $b_2 = c_2$, the burn-in of \Cref{lem: weights lower bound} dominates \Cref{lem: weights upper bound}, we get the following bound on the contraction factor
\begin{lemma}\label{lem: contraction factor bound}
    Let
    \begin{align*}
        N_1 &\gtrsim \max\curly{\gamma^4\paren{r + \log(T/\delta)},  \frac{\bfFmin^{-1}}{b_2^2}\frac{1}{T}\sumT \frac{\sigmahw^2 (d_y + r + \log(T/\delta))}{\lambda_{\min}(\Sigmaxh) } },
    \end{align*}
    and $\dist(\hat\Phi, \Phi_\star) \leq \frac{4}{5}b_1 \sqrt{\frac{\bfFmin}{\bfFmax}} \max_t \kappa\mem\paren{\Sigmaxh}^{-1}$ for given constants $b_1,b_2 \in (0,1)$. Then, for step-size satisfying $\eta \leq \frac{1}{(1 + 2b_1 + 2b_2 + (b_1 + b_2)^2))\bfFmax}$, with probability at least $1 - \delta$, we have
    \begin{align*}
        \norm{I_{d_x} - \eta \frac{1}{T}\sum_{t=1}^T\hatFh^\top \hatFh} &\leq \paren{1 - \paren{(1 - \frac{2}{3}b_1^2) - 2\paren{b_1 + b_2 + b_1 b_2}}\eta \bfFmin }.
    \end{align*}
\end{lemma}
We make a couple of qualitative remarks here:
\begin{itemize}
    \item When $b_1, b_2$ are small, and $\eta$ is set to its maximal permitted value, then the contraction rate approaches $\paren{1 - \frac{\bfFmin}{\bfFmax}}$, which is intuitively the best one can hope for by inspecting the contraction factor $I - \frac{\eta}{T}\sumT \hatFh^\top \hatFh$.

    \item Though the above initialization requirement on $\dist(\hat\Phi, \Phi_\star)$ may be relevant during the early iterations, we note that due to the exponential convergence ensured by \Cref{lem: contraction factor bound}, the requirement (i.e.\ $b_1$) can be scaled down exponentially quickly, leaving the dominant barrier the burn-in requirement on $N_1$.
\end{itemize}


The last remaining step is to bound the effect of the orthogonalization factor $R$. We want to upper bound $\|R^{-1}\| = 1/\sigma_{\min}(R)$, and thus it suffices to lower bound $\sigma_{\min}(R)$. By definition, we have
\begin{align*}
    &RR^\top = (R\hat\Phi_+)(R\hat\Phi_+)^\top \\
    =\;& \paren{ \hat\Phi - \frac{\eta}{T}\sum_{t=1}^T \mathopen{}\hat F^{(t)}\mathclose{}^\top \paren{\hat F^{(t)} \hat\Phi - \Fh_\star \Phi_\star} - \frac{\eta}{T}\sum_{t=1}^T \mathopen{}\hat F^{(t)}\mathclose{}^\top \Wh^\top \Xh \paren{\hatSigmaxhNT}^{-1}}  \\
    &\paren{ \hat\Phi - \frac{\eta}{T}\sum_{t=1}^T \mathopen{}\hat F^{(t)}\mathclose{}^\top \paren{\hat F^{(t)} \hat\Phi - \Fh_\star \Phi_\star} - \frac{\eta}{T}\sum_{t=1}^T \mathopen{}\hat F^{(t)}\mathclose{}^\top \Wh^\top \Xh \paren{\hatSigmaxhNT}^{-1}}^\top \\
    \succeq\;&I_r - \underbrace{\Sym\paren{\frac{\eta}{T}\sum_{t=1}^T \mathopen{}\hat F^{(t)}\mathclose{}^\top \paren{\hat F^{(t)} \hat\Phi - \Fh_\star \Phi_\star}\hat\Phi^\top} }_{(a)} 
    - \underbrace{\Sym\paren{\frac{\eta}{T}\sum_{t=1}^T \mathopen{}\hat F^{(t)}\mathclose{}^\top \Wh^\top \Xh \paren{\hatSigmaxhNT}^{-1}\hat\Phi^\top}}_{(b)} \\
    &\quad + \underbrace{\Sym\paren{\frac{\eta^2}{T^2} \mathopen{}\hat F^{(t)}\mathclose{}^\top \paren{\hat F^{(t)} \hat\Phi - \Fh_\star \Phi_\star}^\top \paren{\sum_{t=1}^T \mathopen{}\hat F^{(t)}\mathclose{}^\top \Wh^\top \Xh \paren{\hatSigmaxhNT}^{-1}}^\top }}_{(c)},
\end{align*}
where, besides $\hat\Phi \hat\Phi^\top = I_r$, we have discarded the pd diagonal terms of the expansion. Intuitively, we will show that under appropriate burn-in conditions $\norm{\frac{1}{T} \sumT \mathopen{}\hat F^{(t)}\mathclose{}^\top \paren{\hat F^{(t)} \hat\Phi - \Fh_\star \Phi_\star}} \lesssim \bfFmin$ and $\norm{\frac{1}{T}\sum_{t=1}^T \mathopen{}\hat F^{(t)}\mathclose{}^\top \Wh^\top \Xh \paren{\hatSigmaxhNT}^{-1}} \lesssim \bfFmin$, which will in turn imply
\begin{align*}
    RR^\top &\succcurlyeq (1 - c\eta \bfFmin) I_r
    \implies 1/\sigma_{\min}(R) \lesssim (1 - c\eta \bfFmin)^{-1/2},
\end{align*}
which deflates the effective contraction rate established in \Cref{lem: contraction factor bound} by a square-root. It remains to establish the requisite bounds on the cross-terms.

Focusing on the first cross term (a), we have
\begin{align*}
    &\Sym\paren{\frac{\eta}{T}\sum_{t=1}^T \mathopen{}\hat F^{(t)}\mathclose{}^\top \paren{\hat F^{(t)} \hat\Phi - \Fh_\star \Phi_\star}\hat\Phi^\top} \\
    =\;&\frac{\eta}{T}\sum_{t=1}^T\Sym\paren{\hatFh^\top \paren{\hatFh - \Fhstar \Phi_\star\hat\Phi^\top} } \\
    =\;&\frac{\eta}{T}\sum_{t=1}^T\Sym\paren{\hatFh^\top \paren{\Deltah + \Eh} } \\
    =\;&\frac{\eta}{T}\sum_{t=1}^T\Sym\paren{\paren{\Fhstar\Phi_\star \hat\Phi^\top + \Deltah + \Eh}^\top \paren{\Deltah + \Eh} } \\
    =\;&\frac{\eta}{T}\sum_{t=1}^T
    \Sym\bigg(\hat\Phi \Phi_\star^\top \Fhstar^\top \Deltah + \hat\Phi \Phi_\star^\top \Fhstar^\top \Eh
    \bigg) + 2\Deltah^\top \Deltah + 2 \Eh^\top \Eh + 2\Sym\paren{\Deltah^\top \Eh}.
\end{align*}
Much like for \Cref{lem: weights lower bound}, it suffices to determine bounds on $\dist(\hat\Phi, \Phi_\star)$ and $N_1$ such that the individual terms above are upper bounded by a desired constant factor of $\bfFmin$. For given $c_1, c_2 \in (0,1)$, if the following hold:
\begin{align*}
    \dist(\hat\Phi, \Phi_\star) &\leq c_1 \sqrt{\frac{\bfFmin}{\bfFmax}} \max_t \kappa\mem\paren{\Sigmaxh}^{-1} \\
    N_1 &\gtrsim \frac{\bfFmin^{-1}}{c_2^2}\frac{1}{T}\sumT \frac{\sigmahw^2 (d_y + r + \log(1/\delta))}{\lambda_{\min}(\Sigmaxh) },
\end{align*}
then we have
\begin{align*}
    \lambda_{\max}\paren{\Sym\paren{\frac{\eta}{T}\sum_{t=1}^T \mathopen{}\hat F^{(t)}\mathclose{}^\top \paren{\hat F^{(t)} \hat\Phi - \Fh_\star \Phi_\star}\hat\Phi^\top}} &\leq 2\paren{c_1 + c_2 + 2\paren{c_1 + c_2}^2} \eta \bfFmin.
\end{align*}
Now we notice the second cross term (b) is precisely the noise term considered in \Cref{appdx: noise term bound}, and thus given the burn-in
\begin{align*}
    N_1 &\gtrsim \max\curly{\gamma^4 \paren{r + \log(T/\delta)}, \max_t \frac{\sigmahw^2}{C^2\snorm{\Fhstar}^2\lambda_{\min}(\Sigmaxh)} \paren{d_y + r + \log(T/\delta)}}, \\
    N_2 &\gtrsim \gamma^4 \paren{d_x + \log(T/\delta)},
\end{align*}
and $\dist(\hat\Phi, \Phi_\star) \leq \max_t \frac{2}{5}C \kappa\mem\paren{\Sigmaxh}^{-1}$, for fixed $C > 0$, with probability at least $1 - \delta$,
\begin{align*}
    \norm{\frac{1}{T}\sum_{t=1}^T \mathopen{}\hat F^{(t)}\mathclose{}^\top \Whtwo^\top \Xhtwo \paren{\hat\Sigma^{(t)}_{x, \calN_2}}^{-1}} &\lesssim
    \sigmaavg (1+C) \sqrt{\frac{d_x + \log(T/\delta)}{TN_2} \log\paren{\frac{d_x}{\delta}}},
\end{align*}
where we recall $\sigmaavg := \sqrt{\frac{1}{T}\sumT \frac{\sigmahw^2 \snorm{\Fhstar}^2}{\lambda_{\min}(\Sigmaxh)}}$ is the \emph{task-averaged} noise-level. Setting $C = 2b_1$ (where $b_1 \leq 1/2$ is determined in \Cref{lem: weights lower bound}), the requirement on $\dist(\hat\Phi, \Phi_\star)$ becomes redundant and $(1+C) \leq 2$. Then, we may invert the RHS of the above inequality for $c_3 \bfFmin$ to yield a burn-in on $TN_2$: setting
\begin{align*}
    TN_2 \gtrsim \frac{\bfFmin^{-1}}{c_3^2} \sigmaavg^2 (d_x + \log(T/\delta)) \log\paren{\frac{d_x}{\delta}},
\end{align*}
then with probability at least $1 - \delta$
\begin{align*}
    \lambda_{\max}\paren{\Sym\paren{\frac{\eta}{T}\sum_{t=1}^T \mathopen{}\hat F^{(t)}\mathclose{}^\top \Wh^\top \Xh \paren{\hatSigmaxhNT}^{-1}\hat\Phi^\top}} &\leq c_3 \eta \bfFmin.
\end{align*}
We note that the third and last cross term (c) is bounded by the product of our bounds on the first two cross terms, which under our conditions are both bounded by 1. Piecing these bounds together yields the following bound on the orthogonalization factor.

\begin{lemma}\label{lem: orthogonalization factor bound}
    Let the following burn-in conditions hold:
    \begin{align*}
        \dist(\hat\Phi, \Phi_\star) &\leq \frac{4}{5} c_1 \sqrt{\frac{\bfFmin}{\bfFmax}} \max_t \kappa\mem\paren{\Sigmaxh}^{-1} \\
        N_1 &\gtrsim \max\bigg \{\gamma^4 \paren{r + \log(T/\delta)}, \max_t \frac{\sigmahw^2}{c_1^2\snorm{\Fhstar}^2\lambda_{\min}(\Sigmaxh)} \paren{d_y + r + \log(T/\delta)}, \\
        &\qquad \qquad \qquad \qquad \qquad \qquad \frac{\bfFmin^{-1}}{c_2^2}\frac{1}{T}\sumT \frac{\sigmahw^2}{\lambda_{\min}(\Sigmaxh) }  (d_y + r + \log(T/\delta))\bigg\}, \\
        N_2 &\gtrsim \max\curly{\gamma^4 \paren{d_x + \log(T/\delta)},  \frac{\bfFmin^{-1}}{c_3^2} \blue{\frac{\sigmaavg^2}{T}} (d_x + \log(T/\delta)) \log\paren{\frac{d_x}{\delta}}  },
    \end{align*}
    where $c_1, c_2, c_3 \in (0,1/2)$ are fixed constants. Then, given $\eta \leq \frac{1}{(1 + 2c_1 + 2c_2 + (c_1 + c_2)^2))\bfFmax}$, with probability at least $1 - \delta$, we have the following bound on the orthogonalization factor $R$:
    \begin{align*}
        \norm{R^{-1}} &\leq \paren{1 - \paren{c + c_3 + c\cdot c_3 }\eta \bfFmin}^{-1/2},
    \end{align*}
    where $c := 2(c_1 + c_2 + 2(c_1 + c_2)^2)$.
\end{lemma}

We are now ready to combine \Cref{lem: contraction factor bound} and \Cref{lem: orthogonalization factor bound} to yield the full high-probability descent guarantee for \texttt{DFW}. From those lemmas, we have the free parameters $b_1, b_2$ and $c_1, c_2, c_3$ that trade-off between the burn-in and the contraction factor. Recall the upper bound on our contraction factor scales as
\begin{align}
     &\norm{R^{-1}} \cdot \norm{I_{d_x} - \eta \frac{1}{T}\sum_{t=1}^T\hatFh^\top \hatFh} \nonumber \\
     \leq&\; \frac{1 - \paren{(1 - \frac{2}{3}b_1^2) - 2\paren{b_1 + b_2 + b_1 b_2}}\eta \bfFmin }{\sqrt{1 - \paren{c + c_3 + c\cdot c_3 }\eta \bfFmin}}. \label{eq: orthonormalization-adjusted rate}
\end{align}
To simplify the above, it suffices to set $b_1 = c_1$, $b_2 = c_2$. Therefore, for sufficiently small $c_1, c_2, c_3$, the factor preceding $\eta \bfFmin$ in the numerator will be larger than that on the denominator, which allows us to upper bound the whole contraction factor as simply the square-root of the numerator, as in \cite{collins2021exploiting}. However, slowing the contraction rate by a square root is qualitatively wasteful, as for sufficiently small $c_1, c_2$, the numerator approaches $1 - \eta \bfFmin$, while the denominator approaches $1$. Therefore, we should expect that the rate should typically not suffer the square-root, captured in the following derivation.
\begin{lemma}\label{lem: avoiding square root rate}
    Given $a_1, a_2, d \in (0,1)$ and $\varepsilon \in (0,1)$, if $\varepsilon \geq a_2/a_1$, then the following holds:
    \begin{align*}
        \frac{1 - a_1 d}{\sqrt{1 - a_2 d}} < 1 - (1 - \varepsilon)a_1 d.
    \end{align*}
    Additionally, as long as $\varepsilon \leq 1-\frac{1-\sqrt{1-a_1d}}{a_1 d}$, then $1 - (1-\varepsilon)a_1 d \leq \sqrt{1 - a_1 d}$.
\end{lemma}

\textit{Proof of \Cref{lem: avoiding square root rate}:} squaring the desired inequality and re-arranging some terms, we arrive at
\begin{align*}
    a_2  &\leq \frac{1}{d}\paren{1 - \frac{(1-a_1 d)^2}{(1 - (1-\varepsilon)a_1d)^2}} \\
    &= \frac{1}{d}\paren{1 - \frac{1-a_1 d}{\underbrace{1 - (1-\varepsilon)a_1d}_{< 1}}} \underbrace{\paren{1 + \frac{1-a_1 d}{1 - (1-\varepsilon)a_1d}}}_{> 1}.
\end{align*}
To certify the above inequality, it suffices to lower-bound the RHS. Since $a_1, a_2, d \in (0,1)$, the last factor is at least $1$, such that we have
\begin{align*}
    \frac{1}{d}\paren{1 - \frac{1-a_1 d}{1 - (1-\varepsilon)a_1d}} \paren{1 + \frac{1-a_1 d}{1 - (1-\varepsilon)a_1d}} &>  \frac{1}{d}\paren{1 - \frac{1-a_1 d}{1 - (1-\varepsilon)a_1d}} \\
    &= \frac{1}{d}\frac{(1 - \varepsilon)a_1 d}{1 - (1-\varepsilon)a_1d} \\
    &> \varepsilon a_1.
\end{align*}
Therefore, $a_2 \leq \varepsilon a_1$ is sufficient for certifying the desired inequality. The latter claim follows by squaring and rearranging terms to yield the quadratic inequality:
\begin{align*}
    (1-\varepsilon)^2 a_1 d - 2(1-\varepsilon) + 1 \leq 0,
\end{align*}
Setting $\lambda := 1-\varepsilon$, the solution interval is $\lambda \in \paren{\frac{1-\sqrt{1-a_1 d}}{a_1 d}, \frac{1 + \sqrt{1-a_1 d}}{a_1 d}}$. The upper limit is redundant as it exceeds 1 and $\varepsilon \in (0,1)$, leaving the lower limit as the condition on $\varepsilon$ proposed in the lemma.

$\qedhere$

To operationalize \Cref{lem: avoiding square root rate}, taking $d := \eta \bfFmin$, $a_1$ as the constants on the numerator of \eqref{eq: orthonormalization-adjusted rate}, and $a_2$ as the constants on the denominator, as long as we choose $c_1, c_2, c_3$ such that $a_2$ is smaller than a given fraction of $a_1$, then the descent rate is essentially preserved after accounting for the orthonormalization factor. The latter claim captures that avoiding the square-root is ``worth it'' as long as $a_1 d$ is not too close to $1$, which is usually satisfied considering $d = \eta \bfFmin$ is at largest $\frac{1}{(1 + 2c_1 + 2c_2 + (c_1 + c_2)^2))}\frac{\bfFmin}{\bfFmax}$ by \Cref{lem: orthogonalization factor bound}. Therefore, choosing $a_3 := (1 + 2c_1 + 2c_2 + (c_1 + c_2)^2))$, we can ensure our choice of $\varepsilon$ is valid independent of $\bfFmin, \bfFmax$ by checking $\varepsilon \leq 1 - \frac{1 - \sqrt{1 - a_1/a_3}}{a_1/a_3}$. Since this implies the choice of $\varepsilon$ must satisfy a lower and upper bound $\varepsilon \in \paren{\frac{a_2}{a_1}, \; 1 - \frac{1 - \sqrt{1 - a_1/a_3}}{a_1/a_3} }$ in terms of a function of our universal constants in $a_1, a_2, a_3$ in order to fulfill the improved convergence rate, we must do our due diligence and instantiate choices of universal constants $c_1,c_2, c_3$ to certify that this acceleration holds universally, independent of problem-dependent parameters.

We note that $c_1, c_2$ also controls the upper bound on $\eta$ in \Cref{lem: orthogonalization factor bound}, with smaller $c_1, c_2$ leading to an upper bound approaching $\eta \leq 1/\bfFmax$. For the purposes of this work, it suffices to set $c_1 \leftarrow 1/80, c_2 = c_3 \leftarrow 1/100$. Conditioned on the events of \Cref{prop: DFW update noise term final bound}, \Cref{lem: contraction factor bound}, \Cref{lem: orthogonalization factor bound}, then plugging in these constants yield the following:
\begin{align*}
    a_1 &= (1 - \frac{2}{3}c_1^2) - 2\paren{c_1 + c_2 + c_1 c_2} \approx 0.955 \\
    a_2 &= c + c_3 + c\cdot c_3 \approx 0.0575 \\
    a_3 &= 1 + 2c_1 + 2c_2 + (c_1 + c_2)^2) \approx 1.046 \\
    \varepsilon &\in \brac{\frac{a_2}{a_1}, 1 - \frac{1 - \sqrt{1 - a_1/a_3}}{a_1/a_3}} \approx [0.0602, 0.228] \neq \emptyset \\
    \implies& \norm{R^{-1}} \cdot \norm{I_{d_x} - \eta \frac{1}{T}\sum_{t=1}^T\hatFh^\top \hatFh} \leq \paren{1 - (1-\varepsilon) a_1 \eta \bfFmin} \\
    &\leq 1 - 0.897 \eta \bfFmin \qquad \text{setting }\varepsilon = a_2/a_1 \\
    \eta &\leq \frac{1}{a_3} \frac{1}{\bfFmax} \approx 0.956 \frac{1}{\bfFmax}.
\end{align*}

\begin{theorem}[{Full version of Theorem \ref{thm: descent guarantee iid}}, iid]\label{appdx: descent guarantee iid}
    Let \Cref{assumption: subgaussian covariate} hold. Let the following burn-in conditions hold:
    \begin{align*}
        \dist(\hat\Phi, \Phi_\star) &\leq \frac{1}{100} \sqrt{\frac{\bfFmin}{\bfFmax}} \max_t \kappa\mem\paren{\Sigmaxh}^{-1} \\
        N_1 &\gtrsim \max \curly{ \gamma^4 \paren{r + \log(T/\delta)}, \;\; \overline{\sigma}_{\bfF}^2\paren{d_y + r + \log(T/\delta)}} , \\
        N_2 &\gtrsim \max\curly{\gamma^4 \paren{d_x + \log(T/\delta)},\;\;  \bfFmin^{-1} \blue{\frac{\sigmaavg^2}{T}} (d_x + \log(T/\delta)) \log\paren{\frac{d_x}{\delta}}  },
    \end{align*}
    where $\overline{\sigma}_{\bfF}^2 := \max\Big\{\max_t \frac{\sigmahw^2}{\snorm{\Fhstar}^2\lambda_{\min}(\Sigmaxh)},  \frac{1}{T}\sumT \frac{\sigmahw^2}{\bfFmin\lambda_{\min}(\Sigmaxh) }\Big\}$. Then, given step-size satisfying $\eta \leq 0.956 \bfFmax^{-1}$, running an iteration of DFW yields an updated representation $\hat\Phi_+$ that satisfies with probability at least $1-\delta$:
    \begin{align*}
        \dist(\hat\Phi_+, \Phi_\star) &\leq \paren{1 - 0.897 \eta\bfFmin} \dist(\hat\Phi, \Phi_\star) + C \cdot \sigmaavg \sqrt{\frac{d_x + \log(T/\delta)}{\blue{TN_2}} \log\paren{\frac{d_x}{\delta}}},
    \end{align*}
    where $C > 0$ is a universal constant and $\sigmaavg := \sqrt{\frac{1}{T}\sumT \frac{\sigmahw^2 \snorm{\Fhstar}^2}{\lambda_{\min}(\Sigmaxh)}}$ is the task-averaged noise level.
\end{theorem}

\subsection{The Non-IID Setting}\label{appx: non-iid guarantees}

To extend our analysis to the non-iid setting, we first instantiate our covariates as $\beta$-mixing stationary processes \citep{yu1994rates, kuznetsov2017generalization}, recalling \Cref{assumption: subgaussian covariate}:
\begin{assumption}[Geometric mixing]\label{assumption: geometric mixing}
    For each $t \in [T]$, assume the process $\scurly{\xhit}_{t\geq 1}$ is a mean-zero stationary $\beta$-mixing process, with stationary covariance $\Sigmaxh$ and $\betah(k) := \Gammah \muh^k$.
\end{assumption}
We note that exact stationarity is unnecessary as long as the marginal distributions converge to stationarity sufficiently fast; however, we assume exact stationarity for notational convenience. We now invoke the blocking technique on each trajectory, where each trajectory is subsampled into $a$ trajectories of length $m$ (where we assume $a$ divides $N$ for notational convenience), by assigning each $a$-th point to a trajectory. We may then apply the analysis of the iid setting on a deflated dataset of $T \cdot m$ data points \textit{drawn from the respective stationary distributions}.
Now applying \Cref{lem: kuznetsov mixing}, setting $g(\cdot)$ as the indicator function for the burn-in requirements of and the final descent bound of \Cref{appdx: descent guarantee iid}, we have for all $j = 1,\dots, a$.
\begin{align*}
    \abs{\Ex\brac{g\paren{\curly{X^{(t), Nm}_\infty}_{t \in [T]}}} - \Ex\brac{g\curly{X^{(t), NT}_{(j)}}_{t \in [T]}}  } \leq  m \beta(a) \leq \delta'
\end{align*}
Setting $\delta' = \delta/a$ and union bounding over each $j =1,\dots, a$, we may invert $N\beta(a) = \delta$ to find $a = \tau_{\mathsf{mix}}^{(t)} := \paren{\frac{\log(\Gammah N/\delta)}{\log(1/\muh)} \vee 1}$. This yields the final descent guarantee adjusting for mixing:

\begin{theorem}[{Full version of Theorem \ref{thm: descent guarantee iid}}, mixing]\label{thm: descent guarantee, non-iid}
    Let \Cref{assumption: subgaussian covariate} hold. Let the following burn-in conditions hold:
    \begin{align*}
        \dist(\hat\Phi, \Phi_\star) &\leq \frac{1}{100} \sqrt{\frac{\bfFmin}{\bfFmax}} \max_t \kappa\mem\paren{\Sigmaxh}^{-1} \\
        N_1 &\gtrsim \max_t \tau_{\mathsf{mix}}^{(t)} \cdot  \max \curly{ \gamma^4 \paren{r + \log(T/\delta)}, \;\; \overline{\sigma}_{\bfF}^2\paren{d_y + r + \log(T/\delta)}} , \\
        N_2 &\gtrsim \max_t \tau_{\mathsf{mix}}^{(t)} \cdot \max\curly{\gamma^4 \paren{d_x + \log(T/\delta)},\;\;  \bfFmin^{-1} \blue{\frac{\sigmaavg^2}{T}} (d_x + \log(T/\delta)) \log\paren{\frac{d_x}{\delta}}  },
    \end{align*}
    where $\overline{\sigma}_{\bfF}^2 := \max\Big\{\max_t \frac{\sigmahw^2}{\snorm{\Fhstar}^2\lambda_{\min}(\Sigmaxh)},  \frac{1}{T}\sumT \frac{\sigmahw^2}{\bfFmin\lambda_{\min}(\Sigmaxh) }\Big\}$. Then, given step-size satisfying $\eta \leq 0.956 \bfFmax^{-1}$, running an iteration of DFW yields an updated representation $\hat\Phi_+$ that satisfies with probability at least $1-\delta$:
    \begin{align*}
        \dist(\hat\Phi_+, \Phi_\star) &\leq \paren{1 - 0.897 \eta\bfFmin} \dist(\hat\Phi, \Phi_\star) + C \cdot \sigmaavg \sqrt{\frac{d_x + \log(T/\delta)}{\blue{TN_2}} \log\paren{\frac{d_x}{\delta}}},
    \end{align*}
    where $C > 0$ is a universal constant and $\sigmaavg := \sqrt{\frac{1}{T}\sumT \frac{\tau_{\mathsf{mix}}^{(t)} 
 \sigmahw^2 \snorm{\Fhstar}^2}{\lambda_{\min}(\Sigmaxh)}}$ is the \emph{task-averaged} noise level.
\end{theorem}
In short, for geometric mixing processes, our algorithm guarantees hold as in the iid setting, deflated effectively by a $\log(N/\delta)$ factor. In particular, past the burn-in, the effect of mixing on the descent rate is averaged across tasks.


\subsection{Converting to Sample Complexity Bounds}

To highlight the importance of the task scaling $T$ in our descent guarantees, we demonstrate how to convert general descent lemmas to sample complexity guarantees.
\begin{lemma}\label{lem: converting offline dataset to online guarantee}
    For a sequence of positive integers $\scurly{M_k}_{k \geq 1} \subset \N$, define $\scurly{d_k}_{k \geq 1} \subset \R_+$ as a sequence of non-negative real numbers dependent on $\scurly{M_k}$ that satisfy the relation
    \[
    d_{k+1} \leq \rho \cdot d_k + \frac{C}{M_k},
    \]
    for some $\rho \in (0,1)$ and $C > 0$. Let $d_0 = \tau$. Given a positive integer $M$, we may partition $M = \sum_{k=1}^K M_k$, where
    \[
    K := \floor{\frac{1}{2}\log\paren{\frac{2}{1+\rho}}^{-1}\frac{M\tau^2}{C}\paren{\frac{1-\rho}{2}}^3 + 1},
    \]
    such that the following guarantee holds on $d_K$:
    \[
    d_K \leq \tau\sqrt{\frac{2C}{M}\paren{\frac{2}{1-\rho}}^3}.
    \]
\end{lemma}
The proof of \Cref{lem: converting offline dataset to online guarantee} follows by setting each $M_k$ such that $\rho \cdot d_k + \frac{C}{M_k} = \paren{\frac{1+\rho}{2}}d_k$, and setting $K$ as the maximal $K$ such that $\sum_{k=1}^K M_k \leq M$. Evaluating $d_K \leq \tau \paren{\frac{1+\rho}{2}}^K$ yields the result. For convenience, we do not consider burn-in times $M_k \geq M_0 \;\forall k$ or pseudo-linear dependence $\frac{C\mathrm{polylog}(M_k)}{M_k}$. However, these will only lead to inflating $d_K$ by a polylog$(M)$ factor.

In essence, \Cref{lem: converting offline dataset to online guarantee} demonstrates how a fixed offline dataset of size $M$ can be partitioned into independent blocks of increasing size such that the final iterate satisfies an approximate ERM bound scaling as $\frac{1}{\sqrt{M}}$, inflated by a function of the contraction rate $\rho$. Instantiating \Cref{lem: converting offline dataset to online guarantee} with the problem parameters of Theorem~\ref{thm: descent guarantee iid} yields \Cref{cor: approximate ERM}.

\subsubsection{Near-ERM Transfer Learning}\label{appx: ERM transfer learning}

An important consequence of \Cref{lem: converting offline dataset to online guarantee} (thus \Cref{cor: approximate ERM}) is that near-ERM parameter recovery bounds can be extracted. In particular, given some $t \in [T+1]$, for a given representation $\hat\Phi$, and the least squares weights $\hatFh$ computed with respect to some independent dataset of size $NT$,
\begin{align*}
    \norm{\hat M^{(t)} - M_\star^{(t)}}_F^2 &= \norm{\hatFh \hat\Phi -\Fhstar \Phi_\star}^2_F \\
    &= \norm{\hatFh \hat\Phi \bmat{\hat\Phi^\top & \hat\Phi_{\perp}^\top} -\Fhstar \Phi_\star\bmat{\hat\Phi^\top & \hat\Phi_{\perp}^\top}}^2_F \\
    &= \norm{\bmat{\hatFh - \Fhstar \Phi_\star \hat\Phi^\top & - \Fhstar \Phi_\star \hat\Phi_\perp^\top}}_F^2 \\
    &= \norm{\hatFh - \Fhstar \Phi_\star \hat\Phi^\top}_F^2 + \norm{\Fhstar \Phi_\star \hat\Phi_\perp^\top}^2_F \\
    &\leq 2\norm{\Fhstar \Phi_\star  \paren{I_{d_x} - \hat\Phi^\top \hat\Phi} \Xh^\top \Zh\paren{\hat\Sigma^{(t)}_{z, \calN_1}}^{-1}}_F^2\\
    &+ 2\norm{\Wh^\top \Zh \paren{\hat\Sigma^{(t)}_{z, \calN_1}}^{-1}}_F^2 
     + \norm{\Fhstar}^2 \dist(\hat\Phi, \Phi_\star)^2 \\
    &\lesssim \norm{\Fhstar}^2 \kappa\paren{\Sigmaxh}\dist(\hat\Phi, \Phi_\star)^2 + \sigmahw^2 \frac{d_y r + \log(1/\delta)}{\lambda_{\min}(\Sigmaxh) NT} \quad \text{w.p. }\geq 1- \delta,
\end{align*}
where the last line follows from applying \Cref{lem: weight term concentration}. We observe that the parameter error nicely decomposes into a term quadratic in $\dist(\hat\Phi, \Phi_\star)$ and least squares fine-tuning error scaling with $\frac{1}{NT}$. For a fixed dataset of size $NT$, one can crudely set aside $\Theta(N)$ samples for each task, and use the rest of the $\Theta(N)$ samples to compute $\hat\Phi$. Invoking \Cref{cor: approximate ERM} and using the set-aside $\Theta(N)$ samples to compute $\hatFh$ conditioned on $\hat\Phi$, we recover the near-ERM high probability generalization bound on the parameter error
\begin{align*}
    \norm{\hat M^{(t)} - M_\star^{(t)}}_F^2 \leq \tilde O \paren{\|\Fhstar\|^2 \kappa\paren{\Sigmaxh}C(\rho) \frac{\max_t \sigmahw^2 d_x r}{NT} +  \frac{\sigmahw^2  d_y r}{\lambda_{\min}(\Sigmaxh) N}}.
\end{align*}

\section{Case Study: Linear Dynamical Systems}\label{appx: linear systems case-study}

To understand the importance of permitting non-isotropy and sequential dependence in multi-task data, we consider the fundamental setting of linear systems, which has served as a staple testbed for statistical and algorithmic analysis in recent years, since it lends itself to non-trivial yet tractable \textit{continuous} reinforcement learning problems (see e.g., \citep{hu2023toward, krauth2019finite, tu2018least, tu2019gap, recht2019tour, fazel2018global}), as well as (online) statistical learning problems with temporally dependent covariates \citep{abbasi2011online, abbasi2013online, simchowitz2020naive, dean2018regret, dean2020sample, agarwal2019logarithmic, agarwal2019online, ziemann2022policy, ziemann2022regret, lee2023fundamental, simchowitz2018learning} (see \citep{tsiamis2022learning} for a tutorial and literature review). In particular for our purposes, the dependence of contiguous covariates in a linear system is intricately connected to its \textit{stability properties} \citep{simchowitz2018learning, jedra2020finite, tu2022learning}, such that we may instantiate the guarantees of \texttt{DFW} for non-iid data in an interpretable manner.

The standard state-space linear system set-up admits the form
\begin{align}
    \begin{split} \label{eq: linear system}
        &s[t+1] = A^{(h)} s[t] + B^{(h)} u[t] + w[t]\\
        &w[t] \iidsim \calN(0, \Sigma^{(h)}_w), \;\;s[0] \sim \calN(0, \Sigma^{(h)}_0),
    \end{split}
\end{align}
where we preemptively index possibly task-specific quantities. We consider the following two common linear system settings: system identification and imitation learning.

\subsection{Linear System Identification}\label{appx: linear sysID guarantees}

In linear system identification, the aim is to estimate the system matrices $(A^{(h)}, B^{(h)})$ given state and input measurements $s_t,u_t$. In particular, we may cast the sysID problem as the following regression:
\[
s[t+1] = \bmat{A^{(h)} & B^{(h)}} \bmat{s[t] \\ u[t]} + w[t].
\]
It is customary to consider exploratory signals that are iid zero-mean Gaussian random vectors $u[t] \iidsim \calN(0, \Sigma^{(h)}_u)$ \citep{simchowitz2018learning, simchowitz2020naive, tsiamis2022learning}. In the stable system case, $\rho(A^{(h)}) < 1$, we can therefore evaluate the covariance of the \textit{stationary} distribution of states $s[t]$ induced by exploratory signal $u[t] \iidsim \calN(0, \Sigma^{(h)}_u)$ by plugging in \eqref{eq: linear system} into the following equation
\begin{align*}
\Ex_{u,w}[s[t]s[t]^\top] &= \Ex_{u,w}\brac{s[t+1]s[t+1]^\top} \\
&= A^{(h)}\Ex_{u,w}\brac{s[t]s[t]^\top} {A^{(h)}}^\top + B^{(h)}\Ex\brac{u[t]u[t]^\top} {B^{(h)}}^\top + \Ex\brac{w[t]w[t]^\top} \\
&= A^{(h)}\Ex_{u,w}\brac{s[t]s[t]^\top} {A^{(h)}}^\top + B^{(h)}\Sigma^{(h)}_u {B^{(h)}}^\top + \Sigma^{(h)}_w
\end{align*}
Therefore, evaluating the stationary state covariance $\Sigma^{(h)}_s:= \Ex\brac{s[\infty]s[\infty]^\top}$ amounts to solving the Discrete Lyapunov Equation (\texttt{dlyap}):
\[
\Sigma_s^{(h)} := A^{(h)}\Sigma_s^{(h)} {A^{(h)}}^\top + B^{(h)}\Sigma^{(h)}_u {B^{(h)}}^\top + \Sigma^{(h)}_w.
\]
In the notation introduced earlier in the paper, casting $y[t] \leftarrow s[t+1]$, $x[t] \leftarrow \bmat{s[t] \\ u[t]}$, $M_\star^{(h)} \leftarrow \bmat{A^{(h)} & B^{(h)}}$, we may instantiate multi-task linear system identification as a non-iid, non-isotropic linear operator recovery problem.
\begin{definition}\label{def: sysID init condition}
    Let the initial state covariance be the stationary covariance $\Sigma_0^{(h)} = \Sigma_s^{(h)}$, such that the covariance of the marginal covariate distribution satisfies
    \begin{align*}
        \Sigma_x^{(h)} := \Ex\brac{x[t]x[t]^\top} = \bmat{\Sigma_s^{(h)} & 0 \\ 0 & \Sigma_u^{(h)}}, \;\text{for all }t \geq 0.
    \end{align*}
\end{definition}
We make the above standard definition for the initial state distribution for convenience, as it ensures the marginal distributions of each state are identical. We note, however, given a different initial state distribution, the marginal state distribution converges exponentially quickly to stationarity, thus accumulating only a negligible factor to the final rates. We then make the following system assumptions to instantiate our representation learning guarantees.
\begin{assumption}\label{assumption: linear sysID}
    We assume that for any task $h$ the following hold:
    \begin{enumerate}
        \item The operators share a rowspace $M_\star^{(h)} := \bmat{A^{(h)} & B^{(h)}} = F_\star^{(h)} \Phi_\star$, $F_\star^{(h)} \in \R^{d_s \times r}$, $\Phi_\star \in \R^{r \times (d_s + d_u)}$.

        \item The state matrices have uniformly bounded spectral radii $\rho(A^{(h)}) < \mu < 1$. Subsequently, we assume there exists a constant $\Gamma' > 0$ that satisfies
        \[
        \|{A^{(h)}}^{k}\|_2 \leq \Gamma' \mu^k, \;\;\text{for all }k \geq 0.
        \]
        The existence of a uniform $\Gamma'$ is guaranteed by Gelfand's Formula \citep{horn2012matrix}, and quantitative bounds may be found in, e.g., \cite{goldenshluger2001nonasymptotic, tu2018least}.
    \end{enumerate}

\end{assumption}

The first assumption is satisfied, for example, when $A^{(h)} = P_\star^{(h)} U_\star$ and $B^{(h)} = Q_\star^{(h)} V_\star$ individually admit low-rank decompositions. The second assumption translates to a quantitative bound on the mixing time of the covariates $x[t]$ by adapting a result from \cite{tu2018least}.
\begin{proposition}[{Adapted from \citet[Prop.\ 3.1]{tu2018least}}]\label{prop: mixing time sysID}
    For each $h$, let the dynamics for the linear system evolve as described in \eqref{eq: linear system}. Let \Cref{assumption: linear sysID} hold with constants $\Gamma', \rho$. Define $\prob_{s[k] \sim \nu_k}\brac{\;\cdot \mid s_0 = s}$ as the conditional distribution of states $s[k]$ given initial condition $s_0 = s$. We have for any $k \geq 0$ and initial state distribution $\nu_0$,
    \begin{align}
        \Ex_{s \sim \nu_0}\brac{\norm{\prob_{s[k]}\brac{\;\cdot \mid s_0 = s} - \prob_{s[k]}}_{\mathrm{tv}}} \leq \frac{\Gamma'}{2}\sqrt{\Ex_{\nu_0}[\norm{s[0]}^2] + \frac{\norm{\Sigma^{-1}}_{\ast}}{1 - \mu^2}} \cdot \mu^k,
    \end{align}
    where $\norm{\cdot}_\ast$ indicates the nuclear norm \citep{horn2012matrix}, and $\Sigma := B^{(h)} \Sigma_u^{(h)} {B^{(h)}}^\top + \Sigma_w^{(h)}$.
\end{proposition}
We note that by the independence of control inputs $u[t]$, we have trivially that the total variation distance between the conditional and marginal distributions of covariates $x[t]$ is the same as that of the states $s[t]$.
\[
\norm{\prob_{s[k]}\brac{\;\cdot \mid s_0 = s} - \prob_{s[k]}}_{\mathrm{tv}} = \norm{\prob_{x[k]}\brac{\;\cdot \mid s_0 = s} - \prob_{x[k]}}_{\mathrm{tv}}
\]
Since by construction the marginal distribution of states is identically $\calN(0, \Sigma_s^{(h)})$, applying \Cref{prop: mixing time sysID} to $s[t], s[t+k]$ for any $t, k$, we get the following quantitative bound on the mixing-time of the covariates $x[t] = \bmat{s[t]^\top & u[t]^\top}^\top$.
\begin{lemma}\label{lem: mixing time bound sysID}
    Following \Cref{def: sysID init condition} and \Cref{assumption: linear sysID}, the covariate process $\curly{x^{(h)}[t]}_{t\geq 0}$ is a mean-zero, stationary, geometrically $\beta$-mixing process with covariance $\Sigma_x^{(h)} = \bmat{\Sigma_s^{(h)} & 0 \\ 0 & \Sigma_u^{(h)}}$, where $\Sigma_s^{(h)} = \mathrm{dlyap}(A^{(h)}, B^{(h)}\Sigma_u^{(h)}B^{(h)} + \Sigma_w^{(h)})$, and mixing-time bounded by
    \begin{align}
        \begin{split}
            &\beta(k) = \Gamma \mu^k, \;\;\text{where} \\
            &\Gamma := \frac{\Gamma'}{2}\sqrt{\trace\paren{\Sigma_s^{(h)}} + \frac{\norm{\Sigma^{-1}}_{\ast}}{1 - \mu^2}}, \;\Sigma := B^{(h)} \Sigma_u^{(h)} {B^{(h)}}^\top + \Sigma_w^{(h)}.
        \end{split}
    \end{align}
\end{lemma}
Thus, instantiating \Cref{lem: mixing time bound sysID} in Theorem \ref{thm: descent guarantee, non-iid} gives us guarantees of \texttt{DFW} applied to multi-task linear system identification.



\subsection{Imitation Learning}\label{appx: imitation learning guarantees}

In linear (state-feedback) imitation learning (IL), the aim is to estimate linear state-feedback controllers $K^{(h)} \in \R^{d_u \times d_x}$ from (noisy) state-input pairs $\curly{(s[t], u[t])}_{t\geq 0}$ induced by unknown expert controllers $K^{(h)}_\star$. In particular, we assume the expert control inputs are generated as
\begin{align*}
    u[t] = K^{(h)}_\star s[t] + z[t], \quad z[t] \iidsim \calN(0, \Sigma_z^{(h)}),
\end{align*}
which we observe lends itself naturally as a linear regression, casting $y[t] \leftarrow u[t]$, $x[t] \leftarrow s[t]$, $M_\star^{(h)} \leftarrow K_\star^{(h)}$. Plugging the expert control inputs into the dynamics \eqref{eq: linear system} yields that the states/covariates evolve as
\begin{align*}
    s[t+1] &= A^{(h)}s[t] + B^{(h)}\paren{K^{(h)}_\star s[t] + z[t]} + w[t] \\
    &= \paren{A^{(h)} + B^{(h)}K_\star^{(h)}} s[t] + Bz[t] + w[t].
\end{align*}
We make the natural assumption that the expert controller $K_\star^{(h)}$ stabilizes the system, i.e.\ the spectral radius of the closed-loop dynamics has spectral radius strictly less than 1: $\rho\paren{A^{(h)} + B^{(h)}K_\star^{(h)}} < 1$. As such, similar to the linear sysID setting, we may plug the above dynamics into the stationarity equation to yield the stationary covariance:
\begin{align*}
    \Ex[s[t]s[t]^\top] &= \Ex\brac{s[t+1]s[t+1]^\top} \\
    &= \paren{A^{(h)} + B^{(h)}K_\star^{(h)}} \Ex[s[t]s[t]^\top] \paren{A^{(h)} + B^{(h)}K_\star^{(h)}}^\top + B^{(h)} \Sigma_z^{(h)} {B^{(h)}}^\top + \Sigma_w^{(h)} \\
    \implies \Sigma_s^{(h)} &= \mathrm{dlyap}\paren{A^{(h)} + B^{(h)}K_\star^{(h)}, B^{(h)} \Sigma_z^{(h)} {B^{(h)}}^\top + \Sigma_w^{(h)}}.
\end{align*}
Analogously to linear sysID, we make the following assumptions.
\begin{assumption}\label{assumption: imitation learning}
    We assume that for any task $h$ the following hold:
    \begin{enumerate}
        \item The initial state covariance is set to the stationary covariance $\Sigma_0^{(h)} = \Sigma_s^{(h)}$, such that the marginal covariate distributions satisfy
        \[
        \Ex\brac{x[t]x[t]^\top} = \Sigma_s^{(h)} =: \Sigma_x^{(h)}, \;\text{for all }t \geq 0.
        \]

        \item The controllers share a rowspace $M_\star^{(h)} \equiv K_\star^{(h)} = \Fhstar \Phi_\star$, $\Fhstar \in \R^{d_u \times r}$, $\Phi_\star \in \R^{r \times d_s}$.

        \item The closed-loop dynamics have uniformly bounded spectral radii $\rho\paren{A^{(h)} + B^{(h)} K_\star^{(h)}} < \mu < 1$. Subsequently, we assume there exists a constant $\Gamma' > 0$ that satisfies
        \[
        \norm{\paren{A^{(h)} + B^{(h)} K_\star^{(h)}}^k}_2 \leq \Gamma' \mu^k.
        \]
        The existence of uniform $\Gamma'$ is guaranteed by Gelfand's Formula \citep{horn2012matrix}.

    \end{enumerate}
\end{assumption}
By using a result almost identical to \Cref{prop: mixing time sysID}, we yield the following quantitative bound on the mixing time of covariates generated by stabilizing expert controllers.
\begin{lemma}\label{lem: mixing time bound IL}
    Following \Cref{assumption: imitation learning}, the covariate process $\curly{x^{(h)}[t]}_{t\geq 0}$ is a mean-zero, stationary, geometrically $\beta$-mixing process with covariance $\Sigma_x^{(h)} = \Sigma_s^{(h)}$, where $\Sigma_s^{(h)} = \mathrm{dlyap}\paren{A^{(h)} + B^{(h)}K_\star^{(h)}, B^{(h)} \Sigma_z^{(h)} {B^{(h)}}^\top + \Sigma_w^{(h)}}$, and mixing-time bounded by
    \begin{align}
        \begin{split}
            &\beta(k) = \Gamma \mu^k, \;\;\text{where} \\
            &\Gamma := \frac{\Gamma'}{2}\sqrt{\trace\paren{\Sigma_s^{(h)}} + \frac{\norm{\Sigma^{-1}}_{\ast}}{1 - \mu^2}}, \;\Sigma := B^{(h)} \Sigma_z^{(h)} {B^{(h)}}^\top + \Sigma_w^{(h)}.
        \end{split}
    \end{align}
\end{lemma}
Thus, instantiating \Cref{lem: mixing time bound sysID} in Theorem \ref{thm: descent guarantee, non-iid} gives us guarantees of \texttt{DFW} applied to multi-task linear imitation learning.

\section{Additional Numerical Experiments and Details}\label{appx: additional experiments}

We present additional numerical experiments to demonstrate the effectiveness of \texttt{DFW} (Algorithm \ref{alg: multi-task alt min descent}) and provide a more detailed explanation of the task-generating process for constructing random operators in linear regression and system identification examples. Furthermore, we introduce an additional setting, imitation learning, to illustrate the advantages of collaborative learning across tasks in learning a linear quadratic regulator by leveraging expert data to compute a shared common representation across all tasks. In this latter setting, we also emphasize the importance of feature whitening when dealing with non-i.i.d. and non-isotropic data.

\begin{itemize}
    \item \textbf{Random rotation:} For all the numerical experiments presented in this paper, the application of a random rotation around the identity is employed for both task-specific weight generation and the initialization of the representation. This random rotation is defined as $R_{\text{rot}} = \exp(\tilde{L})$, where $\tilde{L}=\frac{L-L^\top}{2}$ and $L = \gamma S$. Here, $S$ is a random matrix with entries drawn from a standard normal distribution, $d_l$ is the corresponding dimension of the high-dimensional latent space, and $\gamma$ corresponds to the scale of the rotation. We set $\gamma=0.01$ for generating different task weights and $\gamma=1$ for initializing the representation.

    \item \textbf{Step-sizes:} The step-size $\eta$ used to update the common representation is carefully selected to ensure a fair comparison between  Algorithm \ref{alg: multi-task alt min descent} and the vanilla alternating minimization-descent approach employed in \texttt{FedRep} \cite{collins2021exploiting}. In Figure \ref{fig:linear_regression}, both the single-task and multi-task implementations of Algorithm \ref{alg: multi-task alt min descent} adopt $\eta =7,5 \times 10 ^{-3}$, whereas the vanilla alternating minimization-descent approach uses $\eta = 7.5 \times 10^{-3}$ for a fair comparison. Similarly, in Figure \ref{fig:sysID}, both the single-task and multi-task versions of Algorithm \ref{alg: multi-task alt min descent} use $\eta = 1 \times 10^{-1}$, while the vanilla alternating minimization-descent approach utilizes $\eta = 2 \times 10^{-3}$.
\end{itemize}

\subsection{Linear Regression with IID and Non-isotropic Data}

Continuing our experiments for the linear regression problem, this time with different random linear operators as illustrated in Figure \ref{fig:sysID}, we present the results for an extended range of tasks using Algorithm \ref{alg: multi-task alt min descent} and the  alternating minimization-descent approach (\texttt{FedRep} \cite{collins2021exploiting}). In this analysis, we utilize the same specific parameters as discussed in \S\ref{sec: numerical}. Additionally, we set the step-size $\eta = 7.5 \times 10^{-3}$ for both the single-task and multi-task implementations of Algorithm \ref{alg: multi-task alt min descent}, and $\eta = 7.5 \times 10^{-5}$ for both the single-task and multi-task  alternating minimization-descent.

\begin{figure}[t]
     \centering
     \includegraphics[width=\textwidth]{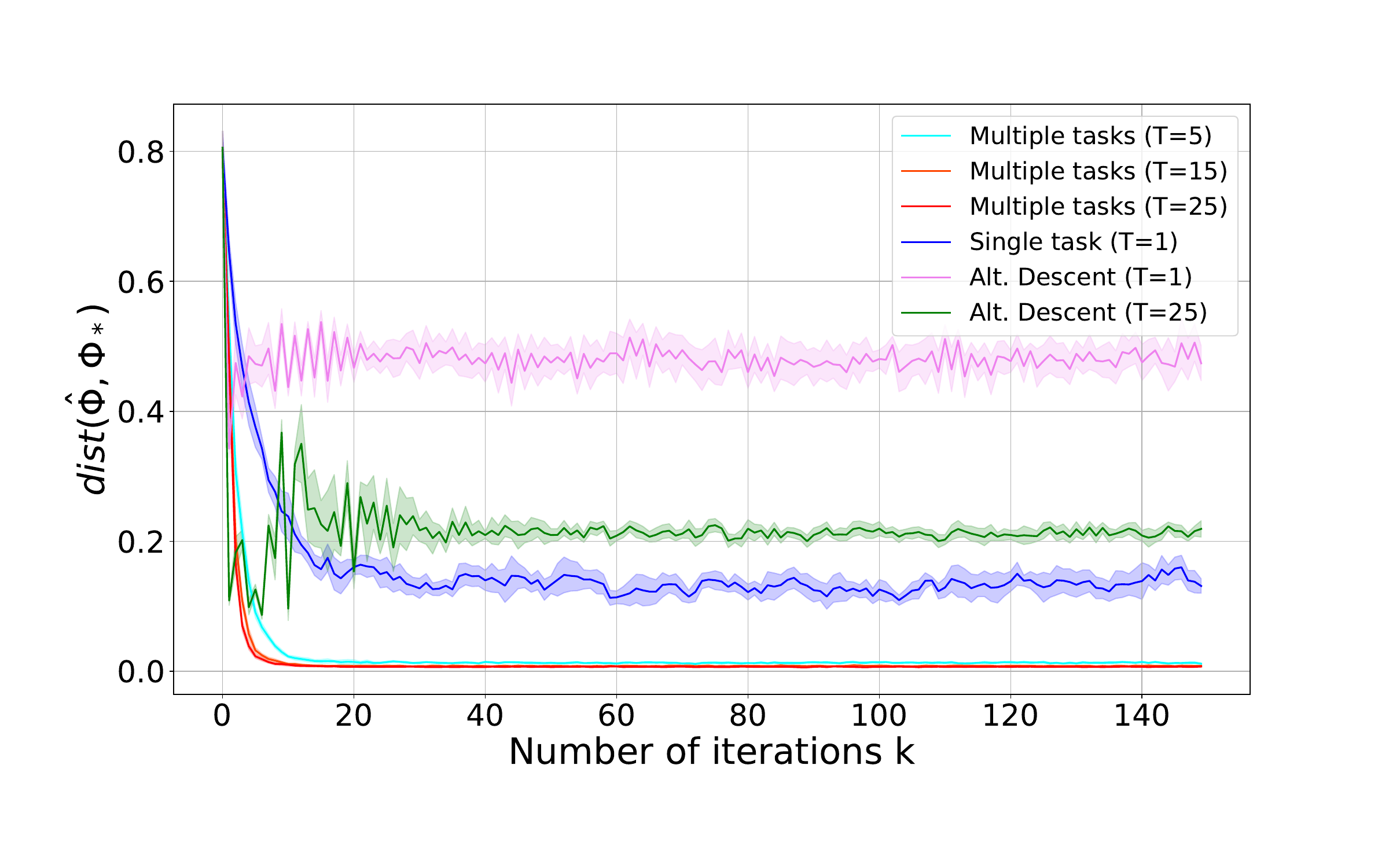}
        \caption{We plot the subspace distance between the current and ground truth representation with respect to the number of iterations, comparing between the single and multiple-task settings of Algorithm \ref{alg: multi-task alt min descent} and the multi-task \texttt{FedRep} for the IID linear regression with random covariance. We observe performance improvement and variance reduction for multi-task \texttt{DFW} as predicted.}
         \label{fig:linear_regression_appendix}
\end{figure}

Figure \ref{fig:linear_regression_appendix} presents a comparison of the performance between Algorithm \ref{alg: multi-task alt min descent} and the vanilla alternating minimization approach in both single and multi-task settings. In line with our theoretical results, the figure demonstrates that as the number of tasks $T$ increases, the error between the current representation and the ground truth representation significantly diminishes. In the specific case of linear regression with iid and non-isotropic data, this figure emphasizes that a small number of tasks ($T=5$), is sufficient to achieve a low error in computing a shared representation across the tasks. Furthermore, the depicted figure reveals that while the multi-task alternating descent algorithm outperforms the single-task case, it is worth noting that this algorithm remains sub-optimal and is unable to surpass the limitation imposed by the presence of bias in the non-isotropic data. Despite its improved performance, the multi-task alternating descent algorithm still encounters challenges in overcoming the inherent noise barrier.   

\subsection{System Identification}

Building upon the results presented in \S\ref{sec: numerical}, we conduct an extended experiment involving a larger range of tasks while maintaining the parameters specified in \S\ref{ss:sysID}. Specifically, we generate distinct random operators different from those utilized to obtain the results illustrated in Figure \ref{fig:sysID}. In this current analysis, we present the outcomes for the expanded range of tasks using Algorithm \ref{alg: multi-task alt min descent} and compare them to the single-task and multi-task vanilla alternating minimization-descent algorithms. The step-size $\eta$ is set to $1 \times 10^{-1}$ for both the single-task and multi-task implementations of Algorithm \ref{alg: multi-task alt min descent}, while for the single-task and multi-task vanilla alternating minimization-descent algorithms, we set $\eta$ to $2 \times 10^{-3}$.

\begin{figure}[t]
     \centering
     \includegraphics[width=\textwidth]{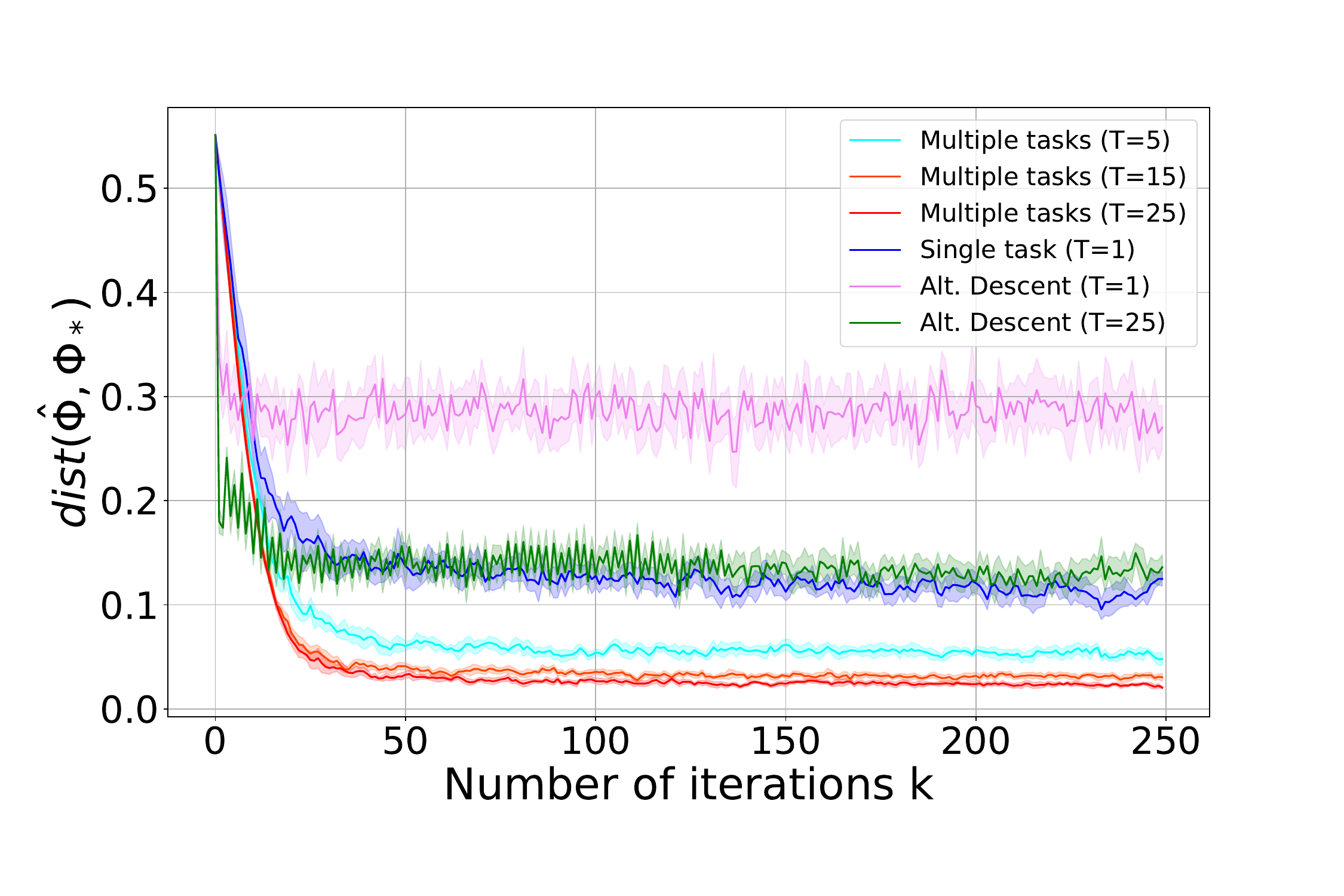}
        \caption{We plot the subspace distance between the current and ground truth representation with respect to the number of iterations, comparing between the single and multiple-task settings of Algorithm \ref{alg: multi-task alt min descent} multi-task \texttt{FedRep} for the linear system identification with random covariance. We observe performance improvement and variance reduction for multi-task \texttt{DFW} as predicted.}
         \label{fig:sysID_appendix}
\end{figure}

In alignment with our main theoretical findings, Figure \ref{fig:sysID_appendix} provides compelling evidence regarding the advantages of the proposed algorithm (Algorithm \ref{alg: multi-task alt min descent}) compared to the vanilla alternating descent approach when computing a shared representation for all tasks. Consistent with the trend observed in Figure \ref{fig:linear_regression_appendix} for the linear regression problem, Figure \ref{fig:sysID_appendix} illustrates a significant reduction in the error between the current representation and the ground truth representation as the number of tasks increases. Additionally, it is noteworthy that while the multi-task alternating descent outperforms the single-task scenario, the single-task variant of Algorithm \ref{alg: multi-task alt min descent} achieves even better results. This observation underscores the importance of incorporating de-biasing and feature-whitening techniques when dealing with non-iid and non-isotropic data.

\subsection{Imitation Learning}

Our focus now turns to the problem of learning a linear quadratic regulator (LQR) controller, denoted as $K^{(T+1)} = F^{(T+1)}_\star \Phi_\star$, by imitating the behavior of $T$ expert controllers $K^{(1)}, K^{(2)},\ldots, K^{(T)}$. These controllers share a common low-rank representation and can be decomposed into the form $K^{(t)} = F^{(t)}_\star \Phi_\star $, where $F^{(t)}_{\star}$ represents the task-specific weight and $\Phi_\star$ corresponds to the common representation across all tasks. To achieve this, we exploit Algorithm \ref{alg: multi-task alt min descent} to compute a shared low-rank representation for all tasks by leveraging data obtained from the expert controllers. Within this context, we consider a discrete-time linear time-invariant dynamical system as follows:
\begin{align*}
    x^{(t)}[i+1]=Ax^{(t)}[i] + Bu^{(t)}[i], \;\ i=0,1,\ldots,N-1,
\end{align*}
with $n_x=4$ states and $n_u=4$ inputs, for all $t \in [T+1]$, where $u^{(t)}[i] = K^{(t)}x^{(t)}[i] + z^{(t)}[i]$, with $z^{(t)}[i] \sim \mathcal{N}(0,I_{n_u})$ being the input noise. In our current setting, rather than directly observing the state, we obtain a high-dimensional observation derived from an injective linear function of the state. Specifically, we assume that $y^{(t)}[i] = Gx^{(t)}[i] + w^{(t)}[i]$, where $G \in \mathbb{R}^{25\times 4}$ represents the high-dimensional linear mapping. The injective linear mapping matrix $G$ is generated by applying a \texttt{thin\_svd} operation to a random matrix with values drawn from a normal distribution $\mathcal{N}(0,1)$. This process ensures injectiveness with a high probability. 

For this aforementioned multi-task imitation learning setting, we adopt a scheme in which we gather observations of the form $\{\{y^{(t)}[i],u^{(t)}[i]\}_{i=0}^{N-1}\}_{t=1}^T$ from the initial $T$ expert controllers to learn the controller $K^{(T+1)}$. These observations are obtained by following the dynamics:
\begin{align*}
    y^{(t)}[i] = (\tilde{A} + \tilde{B}\Tilde{K}^{(t)})y[i] + \Tilde{B}z^{(t)}[i] + w^{(t)}[i]
\end{align*}
with $\Tilde{A} = GAG^{\dagger}$, $\Tilde{B}=GB$, $\tilde{K}^{(t)} = K^{(t)}G^{\dagger}$, and process noise $w^{(t)}[i] \sim \mathcal{N}(0,\Sigma_w)$. 

The collection of stabilizing LQR controllers $K^{(1)}, K^{(2)}, \ldots, K^{(T+1)}$ is generated by assigning different cost matrices, namely $R = \frac{1}{4}I_{n_u}$ and $Q^{(t)}=\alpha^{(t)}I_{n_x}$, where $\alpha^{(t)} \in \texttt{logspace}(0,3, H)$. These matrices are then utilized to solve the Discrete Algebraic Ricatti Equation (DARE): $P^{(t)} = A^\top P^{(t)}A^\top + A^\top P^{(t)} B(B^\top P^{(t)}B + R)^{-1}B^\top P^{(t)} A + Q^{(t)}$, and compute $K^{(t)}=-(B^\top P^{(t)}B + R)^{-1}B^\top P^{(t)} A$, for all $t \in [T+1]$. Moreover, the system matrices $A$ and $B$ are randomly generated, with elements drawn from a uniform distribution. The trajectory length $N=75$ remains consistent for all tasks. The shared representation is initialized by applying a random rotation to the true representation, denoted as $\Phi_\star = G^{\dagger}$.

\begin{figure}[t]
     \centering
     \includegraphics[width=\textwidth]{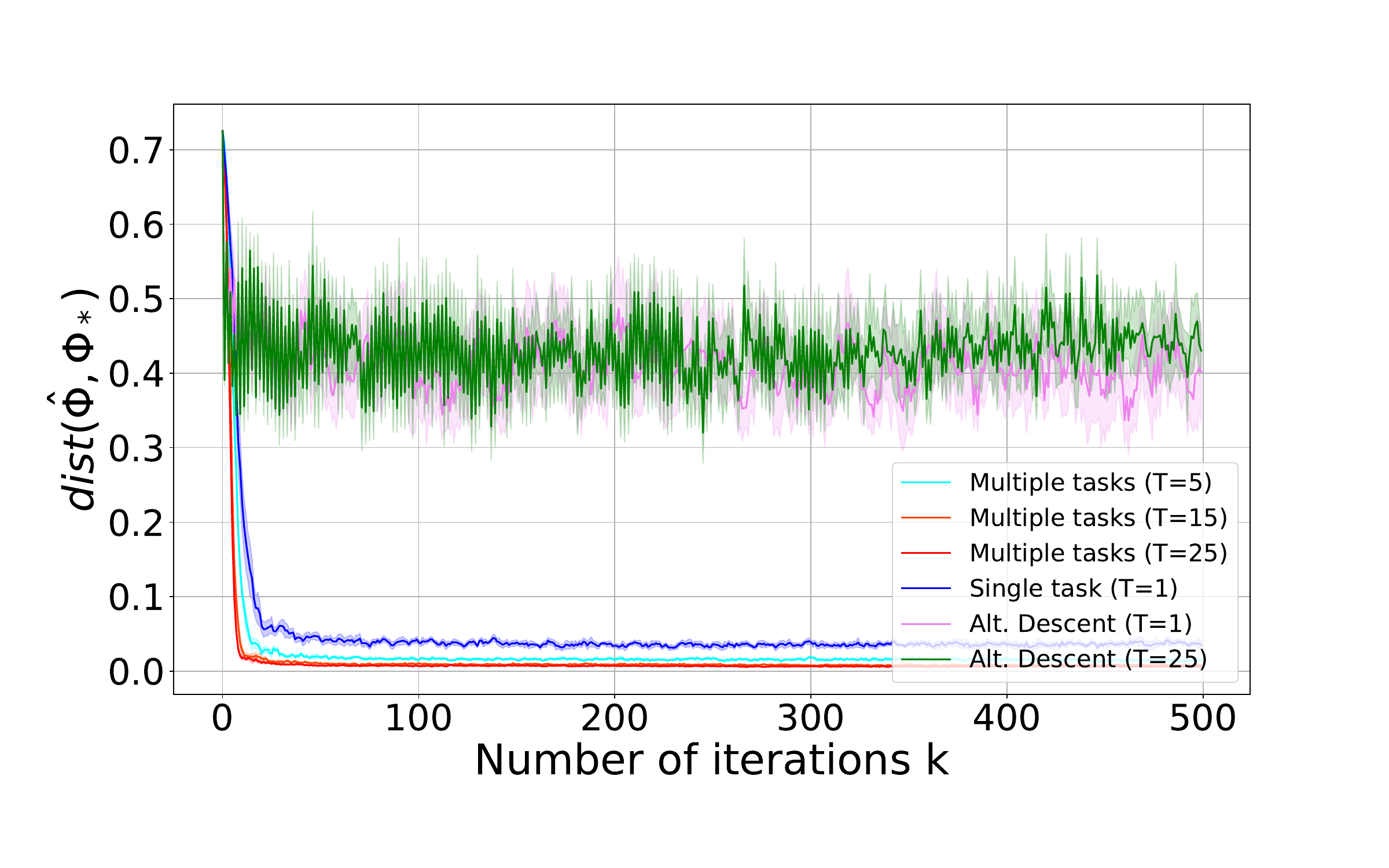}
        \caption{We plot the subspace distance between the current and ground truth representation with respect to the number of iterations, comparing between the single and multiple-task settings of Algorithm \ref{alg: multi-task alt min descent} and the multi-task \texttt{FedRep} for the imitation learning with random covariance. We observe performance improvement and variance reduction for multi-task \texttt{DFW} as predicted.}
         \label{fig:IL_appendix}
\end{figure}

Figure \ref{fig:IL_appendix} presents a comparative analysis between Algorithm \ref{alg: multi-task alt min descent} and the vanilla alternating minimization-descent approach (\texttt{FedRep} in \cite{collins2021exploiting}) for computing a shared representation across linear quadratic regulators. This shared representation is then utilized to derive the learned controller $K^{(T+1)}$ in a few-shot learning manner. Consistent with our theoretical findings and in alignment with the trends observed in Figures \ref{fig:linear_regression_appendix}-\ref{fig:sysID_appendix}, Figure \ref{fig:IL_appendix} demonstrates a substantial reduction in the error between the current representation and the ground truth representation when leveraging data from multiple tasks, compared to the single-task scenario in Algorithm \ref{alg: multi-task alt min descent}. Furthermore, this figure underscores the significance of de-biasing and whitening the feature data in overcoming the bias barrier introduced by non-iid and non-isotropic data. In contrast, the vanilla alternating descent algorithm fails to address this challenge adequately and yields sub-optimal solutions.

\end{document}